\documentclass[9pt,twocolumn,twoside]{article}           
\usepackage[dvips]{graphicx}
\usepackage{amsfonts}
\usepackage{blindtext, rotating}
\usepackage{amssymb}
\usepackage{amsbsy} 
\usepackage{amsmath}
\usepackage[ansinew]{inputenc}
\usepackage{subfigure}
\usepackage{cite}
\usepackage{color}
\usepackage{hyperref}
\usepackage[margin=1.5cm]{geometry}

\newcommand{\A}{\mathbf{A}}

\newcommand{\K}{\mathbf{K}}

\newcommand{\V}{\mathbf{V}}
\newcommand{\W}{\mathbf{W}}

\newcommand{\E}{\mathbf{E}}

\newcommand{\Real}{\mathbb R}

\begin{document}

\title{Dimensionality Reduction via Regression in Hyperspectral Imagery}
\date{}
\author{Valero Laparra, Jes\'us Malo and Gustau Camps-Valls
\thanks{Copyright (c) 2014 IEEE. Personal use of this material is permitted. Permission from IEEE must be obtained for all other users, including reprinting/
republishing this material for advertising or promotional purposes, creating new collective works for resale or redistribution to servers or
lists, or reuse of any copyrighted components of this work in other works. DOI: 10.1109/JSTSP.2015.2417833.}
\thanks{Image Processing Laboratory (IPL), Universitat de Val\`encia, Catedr\'atico A. Escardino - 46980 Paterna, Val\`encia (Spain). E-mail: \{valero.laparra, jesus.malo, gcamps\}@uv.es}
\thanks{This work was partially supported by the Spanish Ministry of Economy and Competitiveness (MINECO) under project TIN2012-38102-C03-01, and under a  EUMETSAT contract.}
}
\pagestyle{myheadings}
\markboth{2015 IEEE. PUBLISHED IN IEEE. DOI: 10.1109/JSTSP.2015.2417833}{Laparra et al., 2015}

\maketitle

\begin{abstract}
This paper introduces a new \emph{unsupervised} method for dimensionality reduction via regression (DRR). The algorithm belongs to the family of {\em invertible transforms} that generalize Principal Component Analysis (PCA) by using curvilinear instead of linear features.
DRR identifies the nonlinear features through multivariate regression to
ensure the reduction in redundancy between the PCA coefficients,
the reduction of the variance of the scores, and the reduction in the reconstruction error.
More importantly, unlike other nonlinear dimensionality reduction methods,
the invertibility, volume-preservation, and straightforward out-of-sample extension, makes DRR
interpretable and easy to apply.
The properties of DRR enable learning a more broader class of data manifolds than the recently proposed Non-linear Principal Components Analysis (NLPCA) and Principal Polynomial Analysis (PPA).
We illustrate the performance of the representation in reducing the dimensionality of remote sensing data.
In particular, we tackle two common problems: processing very high dimensional spectral information such as in hyperspectral image sounding data, and dealing with spatial-spectral image patches of multispectral images. Both settings pose collinearity and ill-determination problems.
Evaluation of the expressive power of the features is assessed in terms of truncation error, estimating atmospheric variables, and surface land cover classification error.
Results show that DRR outperforms linear PCA and recently proposed invertible extensions based on neural
networks (NLPCA) and univariate regressions (PPA).
\end{abstract}

\section{Introduction}\label{sec:intro}
In the last decades, the technological evolution of optical sensors
has provided remote sensing analysts with rich spatial, spectral,
and temporal information. In particular, the increase in spectral
resolution of hyperspectral sensors in general, and of infrared sounders in particular,
opens the doors to new application domains and poses new methodological challenges in data
analysis. The distinct highly-resolved spectra offered by hyperspectral images (HSI)
allow us to characterize land-cover classes with unprecedented accuracy.
For instance, hyperspectral instruments such as NASA's Airborne Visible Infra-Red
Imaging Spectrometer (AVIRIS) covers the wavelength region from 0.4 to
2.5$\mu$m using more than 200 spectral channels, at a nominal
spectral resolution of 10 nm. The MetOp/IASI infrared sounder poses even
more complex image processing problems, as it acquires more than 8000 channels per iFOV.
Actually, such improvements in spectral resolution have called for advances
in signal processing and exploitation algorithms capable of summarizing the information
content in as few components as possible~\cite{CampsValls05,Plaza09,Camps11,Camps-Valls14}.

In addition to its eventual high dimensionality, the complex
interaction between radiation, atmosphere, and objects in the surface leads to
irradiance manifolds which consist of non-aligned clusters that may change nonlinearly in different
acquisition conditions~\cite{Devis13,Laparra12a}.
Fortunately, it has been shown that, given the spatial-spectral smoothness of the signal,
the intrinsic dimensionality of the data is small, and this can be used both for
efficient signal coding~\cite{Penna07,Camps11}, and for knowledge extraction from
a reduced set of features~\cite{Jimenez13,Arenas13}. The high dimensionality problem is not only
affecting the hyperspectral data: very often, multispectral data processing applications involve
using spatial, multi-temporal or multi-angular features that are combined with the spectral features~\cite{Fau08b,Tui09b}. In such cases, the representation space becomes more redundant and pose challenging problems of collinearity for the
algorithms. In both cases, the key in coding, classification, and in bio-geo-physical parameter retrieval applications reduces to finding
the appropriate set of features, that should be necessarily flexible and nonlinear.

In order to find these features, in recent years a number of feature extraction and dimensionality reduction methods have been presented. Most of them are based on nonlinear functions to allow describing data manifolds that exhibit nonlinear relations (see~\cite{Lee07} for a comprehensive review).
Approaches range from local methods~\cite{Tenenbaum2000,Roweis02,Verbeek02,Teh03,Brand03}, kernel-based and spectral decompositions~\cite{Roweis00,Scholkopf98,Weinberger04,Arenas13}, neural networks~\cite{Kramer91,Hinton06,Scholz07}, or projection pursuit formulations~\cite{Huber85,Laparra11}. Despite the theoretical advantages of nonlinear methods, the fact is that classical principal component analysis (PCA)~\cite{Jolliffe02} is still the most widely used dimensionality reduction technique in real remote sensing applications~\cite{CampsValls10eumtgrs,lillesand08,Camps11,Chang08}.
This is mainly because PCA has different properties that make it useful in real examples: it is easy to apply since it involves solving a linear and convex problem, and it has a straightforward out-of-sample extension. Moreover, the PCA transformation is invertible and, as a result, the features extracted can be easily interpreted.

The new dimensionality reduction algorithms that involve nonlinearities rarely fulfill the above properties. Nonlinear models usually have complex formulations, which introduce a number of non-intuitive free parameters. Tuning these parameters implies strong assumptions about the manifold characteristics (e.g. local Gaussianity or special symmetries), or a high computational cost training. This complexity reduces the applicability of nonlinear feature extraction to specific data, i.e. the performance of these methods do not significantly improve that of PCA on many remote sensing problems~\cite{Camps11,CampsValls10eumtgrs,Arenas13}. Moreover, these methods have problems to obtain out-of-sample predictions, which is mandatory in most of the real applications. Another critical point is that the transform involved by the nonlinear models is hard to interpret. This problem could be alleviated if the methods were invertible since then one could get the data back to the input domain (where units are meaningful) and understand the results therein. Invertibility allows to characterize the transformed domain, and to evaluate its quality.
However, invertibility is scarcely achieved in the manifold learning literature.
For instance,
spectral and kernel methods involve \emph{implicit} mappings between the original and the curvilinear coordinates, but these \emph{implicit} features are not easily invertible nor interpretable \cite{Honeine2011a}.

The desirable properties of PCA are straightforward in methods that find projections onto \emph{explicit} features in the input domain.
These \emph{explicit} features may be either straight lines or curves.
This family of projection methods may be understood as a generalization of linear transforms extending linear components to curvilinear components. This family ranges between two extreme cases:
(1) {\bf rigid} approaches where features are straight lines in the input space (e.g. conventional PCA, Independent Components Analysis -ICA- \cite{Hyvarinen2001}), and
(2) {\bf flexible} non-parametric techniques that closely follow the data, such as
Self-Organizing Maps (SOM)~\cite{Kohonen82}, or the related Sequential Principal Curves Analysis (SPCA)~\cite{Laparra12a,Laparra14b}.
This family is discussed in Section~\ref{related} below. Both extreme cases are undesirable because of different reasons:
limited performance (in too rigid methods), and complex tuning of free parameters and/or unaffordable computational cost (in too flexible methods).
In this \emph{projection-onto-explicit-features} context, autoencoders such as Nonlinear-PCA (NLPCA)~\cite{Scholz07}, and approaches based on fitting functional curves, such as Principal Polynomial Analysis (PPA)~\cite{Laparra12b,Laparra14a}, represent convenient intermediate points between the extreme cases in the family. Note that these methods have shown better performance than PCA on a variety of real data~\cite{Laparra14a,Scholz12}. Actually, in the case of PPA, it is theoretically ensured to obtain better results than PCA. The method proposed here, \emph{Dimensionality Reduction via Regression} (DRR), represents a qualitative step towards the flexible end in this family because of the multivariate nature of the regression (as opposed to the univariate regressions done in PPA) while keeping the convenient properties of PPA and PCA which make it suitable for practical high dimensional problems (as opposed to SPCA and SOM). Therefore, it extends the applicability of PPA to more general manifolds, such as those encountered in remote sensing data analysis.

Following the taxonomy in~\cite{Laparra14a} these three methods (NLPCA, PPA and DRR) could be included in the \emph{Principal Curves Analysis} framework \cite{Has84}. This framework includes both parametric (fitting analytic curves)~\cite{Jolliffe02,Donnell94,Besse95}, and non-parametric ~\cite{Einbeck05,Einbeck10,Ozertem11,Laparra12a,Laparra14b} methods. NLPCA, PPA and DRR exploit the idea behind this framework to define generalizations of PCA of \emph{controlled} flexibility.

Preliminary results of DRR were presented in \cite{Laparra14c}. Here we extend the theoretical analysis of the method and the experimental confirmation of the performance in hyperspectral images.
The remainder of the paper is organized as follows.
Section~\ref{related} reviews the properties and shortcomings of the \emph{projection-onto-explicit-features family
pointing out the qualitative advantages of the proposed DRR.}
Section~\ref{the_math_details} introduces the mathematical details of DRR.
We describe the DRR transform and the key differences with PPA.
We derive an explicit expression for the inverse and we prove the volume preservation property of DRR.
The theoretical properties of DRR are demonstrated and illustrated in controlled toy examples of different complexity.
In Section~\ref{experiments}, we address two important high dimensional problems in remote sensing:
the estimation of atmospheric state vectors from Infrared Atmospheric Sounding Interferometer (IASI) hyperspectral sounding data, and the dimensionality reduction and classification of spatio-spectral Landsat image patches.
In the experiments, DRR is compared with conventional PCA~\cite{Jolliffe02}, and with recent fast nonlinear generalizations
that belong to the same class of invertible transforms, PPA~\cite{Laparra12b,Laparra14a} and NLPCA~\cite{Scholz07}.
Comparisons are made both in terms of reconstruction error and of expressive power of the extracted features.
We end the paper with some concluding remarks in Section~\ref{conclusions}.

\section{From rigid to flexible features}\label{related}

Here we illustrate how DRR represents a step forward with regard to NLPCA and PPA
in the family of projections onto explicit curvilinear features ranging from rigid to flexible extremes.
First, we review the basic details of previous projection methods, and then we
illustrate the advantages of the proposed method in an easy to visualize example.

\subsection{Representative projections onto lines and curves}

Classical techniques such as PCA \cite{Jolliffe02} or ICA \cite{Hyvarinen2001} represent the
\emph{rigid} extreme of the family, where, zero-mean samples $x \in {\mathbb R}^d$ are projected onto $d$ rectilinear
features through the projection matrix, $\V$:
\vspace{-0.2cm}
$$
\alpha = \V \cdot x
$$
where $\alpha_i$ are the Principal Components (PC scores for PCA) or the Independent Components (for ICA), and
the $d$ linear features in the input space are the column vectors (straight directions) in $\V^{-1}$.
These rigid techniques use a single set of global features regardless of the input.

On the contrary, flexible techniques adapt the set of features to the local properties of the input.
Examples include SOM \cite{Kohonen82} where a flexible grid
is adapted to the data and samples can be represented by projections onto
the local axes defined by the edges of the parallelepiped corresponding to the closest node.
Similarly, local-PCA \cite{Leen97} and local-ICA \cite{Karhunen00} project the data
onto local axes corresponding to the closest code vector.
More generally, local-to-global methods integrate these local-linear representations
into a single global curvilinear representation \cite{Malo2006b}.
In particular, using the fact that local eigenvectors are
tangent to first and secondary principal curves \cite{Delicado01},
Sequential Principal Curves Analysis (SPCA) \cite{Laparra12a,Laparra14b}
integrates local PCAs, $\V(x')$, along a sequence of $d$ principal curves to get
a curvilinear representation

\vspace{-0.0cm}
$$
r = \int_{x_0}^{x} D(x') \cdot \V(x') \cdot dx',
$$
where the local metric, $D(x')$, sets the line element along the curves.
SPCA is inverted by taking the lengths, $r_i$, along the sequence of principal curves drawn from the origin, $x_0$.
Similarly to SOM, SPCA assumes a grid of curves adapted to the data. However, as opposed to SOM, SPCA does not
learn the whole grid, but only $d$ segments of principal curves per sample.

The above methods identify explicit curves/features that follow the data,
but they are hard to train (e.g. parameters to control their flexibility depend on the problem)
and require many samples to be reliable, which make them hard to use in high-dimensional scenarios.
Other methods have been proposed to generalize the rigid representations by considering curvilinear features
instead of straight lines \cite{Jolliffe02}.
For instance, in NLPCA \cite{Kramer91,Scholz07} an invertible internal representation is computed through a two stage neural network,
$$
r = \W_2 \cdot g_1(\W_1 \cdot x)
$$
where the matrices $\W_i$ represent sets of linear receptive fields, and $g_1$ is a set of fixed point-wise nonlinearities.
The inverse of this autoencoder \cite{Hinton06} can be used to make the curvilinear coordinates explicit.

Fitting general parametric curves in ${\mathbb R}^d$, as done in \cite{Donnell94,Besse95}, is difficult because of the
unconstrained nature of the problem \cite{Jolliffe02,Laparra14a}.
Alternatively, PPA \cite{Laparra14a} follows a deflationary sequence in which a single polynomial
depending on a straight line (univariate fit) is computed at a time.
Specifically, the $i$-th stage of PPA accounts for the $i$-th curvilinear
dimension by using two elements: (1) one-dimensional projection onto the leading vector $e^{(i)}$,
and (2) polynomial prediction of the average at the orthogonal subspace,

\begin{eqnarray}
	\alpha_i &=& e^{(i) \top} \cdot x^{(i-1)} \nonumber \\
    x^{(i)} &=& \E^{(i)}_\bot \cdot x^{(i-1)} - f^{(i)}(\alpha_i)
    \label{PPAapprox}
\end{eqnarray}
where the polynomial prediction, $f^{(i)}(\alpha_i)$, is removed from the data in the orthogonal subspace. Superindices in the above formula represent the stage.
As a result, data at the $i$-th stage is represented by $\alpha_i$ and by the $(d-i)$-dimensional residual that cannot
be predicted from that projection. Prediction using this univariate polynomial is a way
to remove possible nonlinear dependencies between the linear subspace of $e^{(i)}$ and its orthogonal
complement. Despite its convenience, the univariate nature of the fits restricts the kind of dependencies that
can be taken into account since more information about the orthogonal subspace (better predictions)
could be obtained if more dimensions were used in the prediction. Moreover,
using a single parameter, $\alpha_i$, to build the $i$-th polynomial implies that the
$i$-th curvilinear feature has the same shape along the $(i-1)$-th curve.

DRR addresses these limitations by using multivariate instead of univariate regressions in the nonlinear predictions.
As a result, DRR improves energy compaction and extends the applicability of PPA to more general manifolds while keeping its simplicity, which make it suitable in high dimensional problems (as opposed to SPCA and SOM).

\subsection{Qualitative advantages of DRR}

\begin{figure*}[p]
	\centering
    \small
    \setlength{\tabcolsep}{7pt}
    \begin{tabular}{cccccc}
    \hspace{-1.5cm}   & & \small \textbf{Manifold 1}& & \small \textbf{Manifold 2} & \\
    \hspace{-1.5cm}   & & \small (\textbf{easy}: no tilt)& & \small (\textbf{difficult}: tilt)& \\
	   &
    \hspace{-1.5cm} \begin{turn}{90}
\begin{minipage}{3cm}
\begin{center}\small{Input Domain}\end{center}
\end{minipage}
\end{turn} \hspace{-1.5cm}
& \includegraphics[width=3cm]{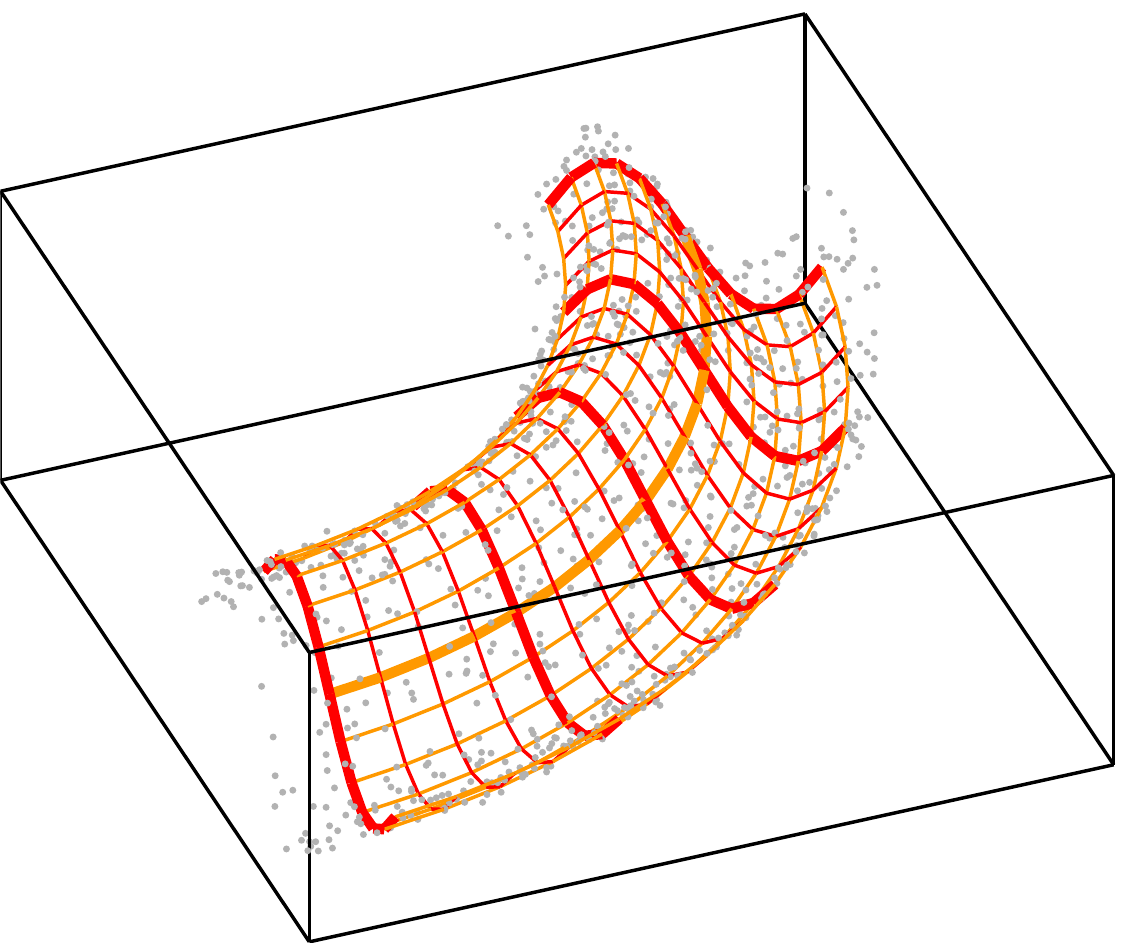} & &
  \includegraphics[width=3cm]{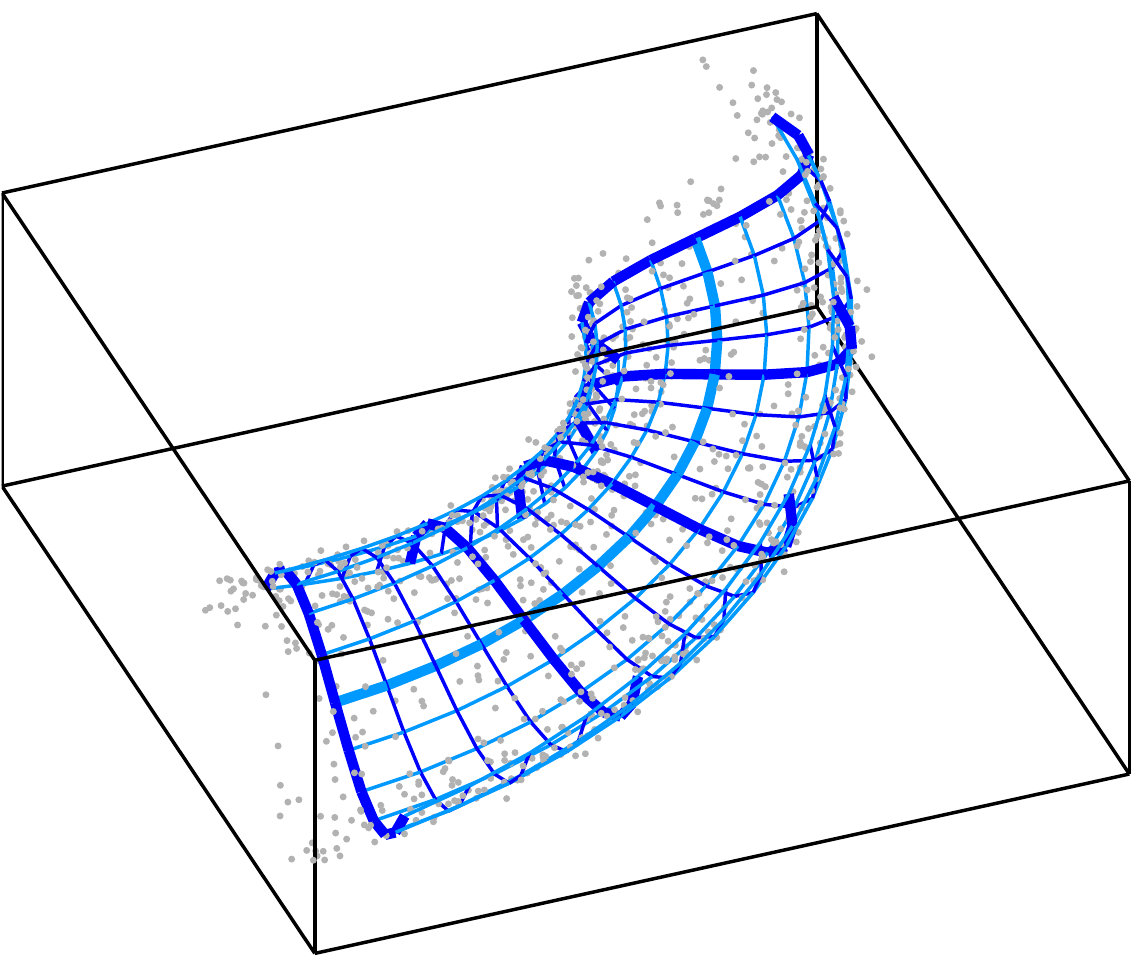} & \\[2mm]
    \end{tabular}
    \begin{tabular}{cccc|c|c}
        \hline \\[2mm]
    \hspace{-1.5cm}  &  \small \textbf{Rigid Extreme} &  &  & & \small \textbf{Flexible Extreme} \\ [2mm]
    \hspace{-1.5cm}  &  \small PCA& \small NLPCA &\small PPA &\small DRR & \small SPCA   \\[-9mm]
\hspace{-1.5cm}\begin{turn}{90}
\begin{minipage}{3cm}
\begin{center}\hspace{-1.5cm} \small{Transform}\end{center}
\end{minipage}
\end{turn} \hspace{-1.5cm}
      &  \includegraphics[width=3cm]{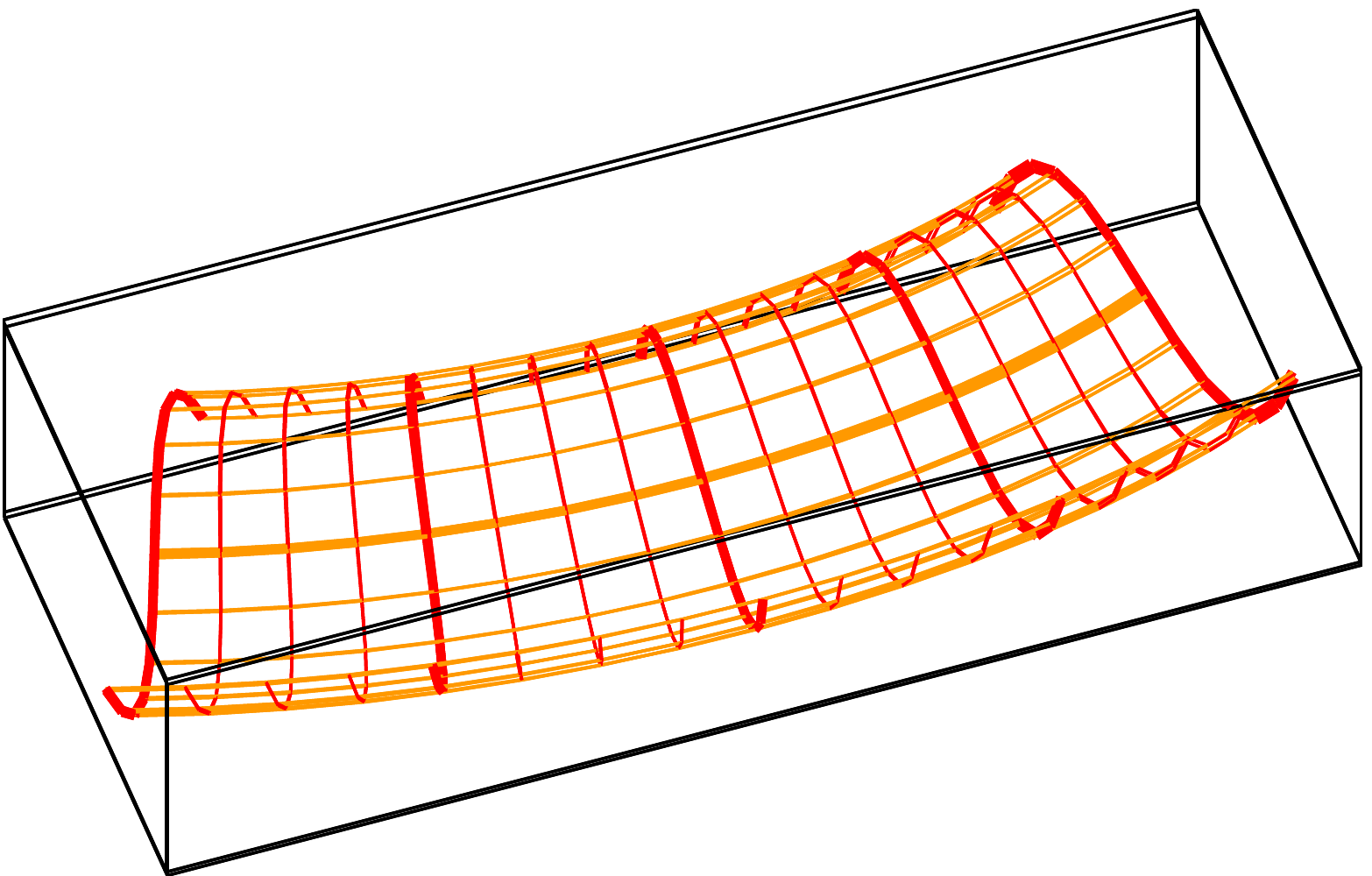} & 
        \includegraphics[width=3cm]{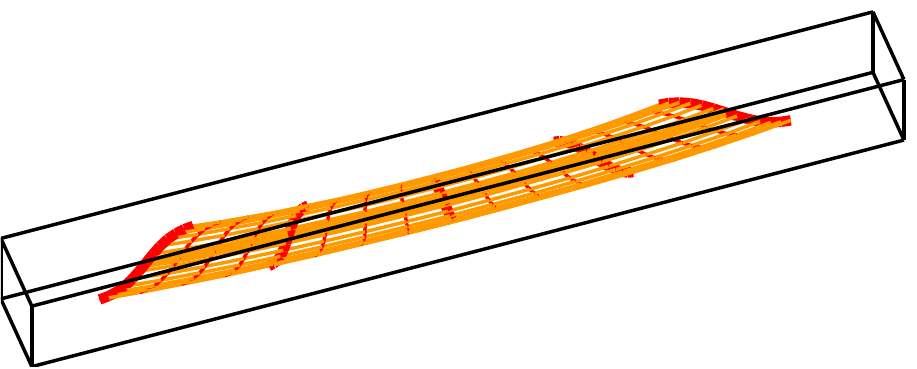} &
        \includegraphics[width=3cm]{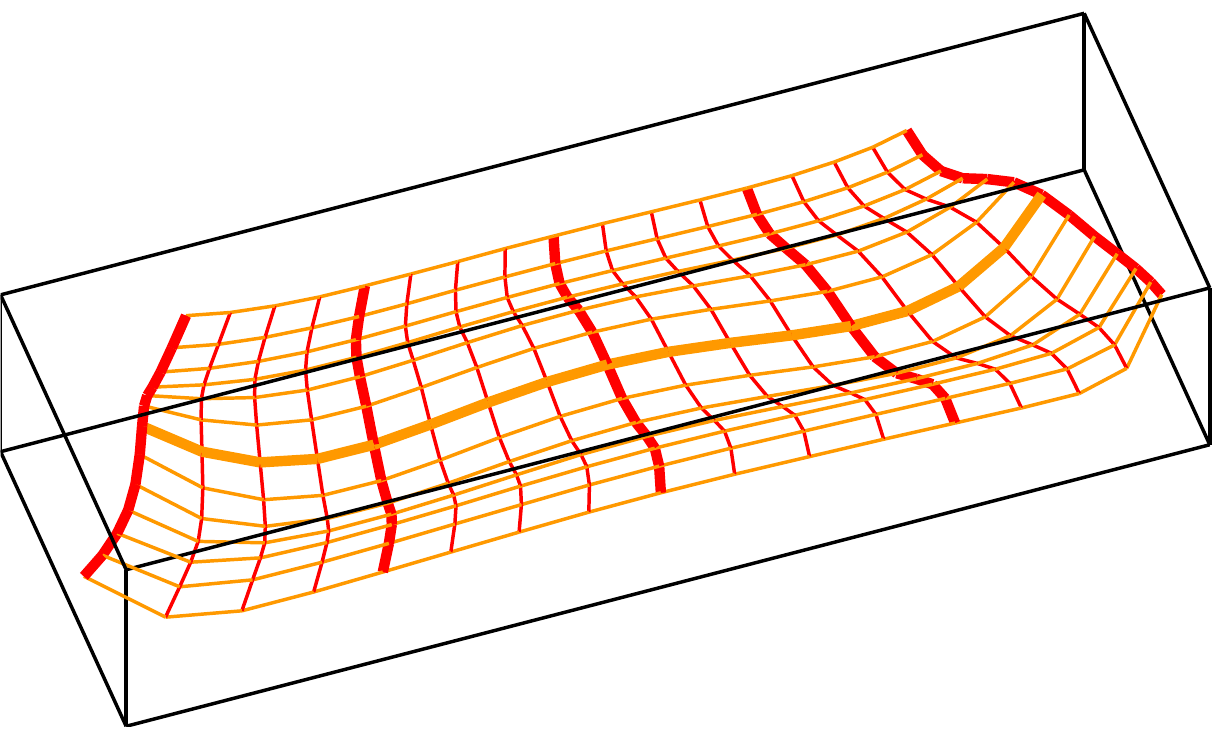} &
        \includegraphics[width=3cm]{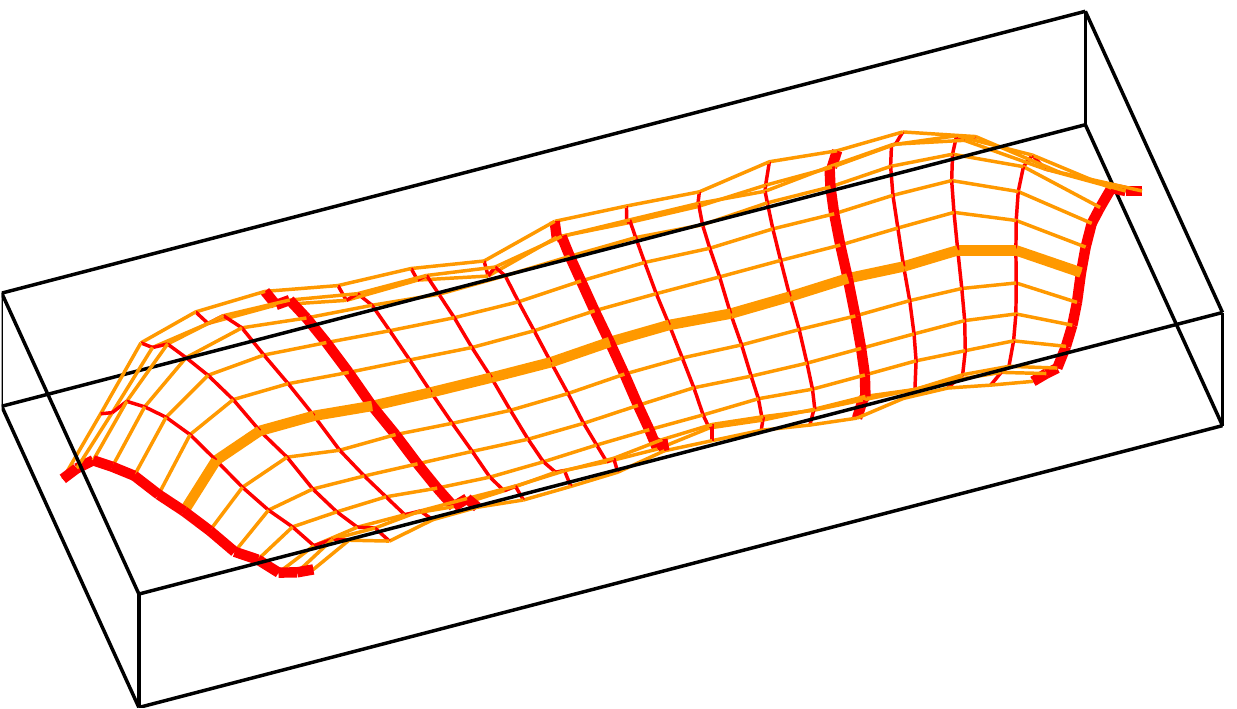} &
        \includegraphics[width=3cm]{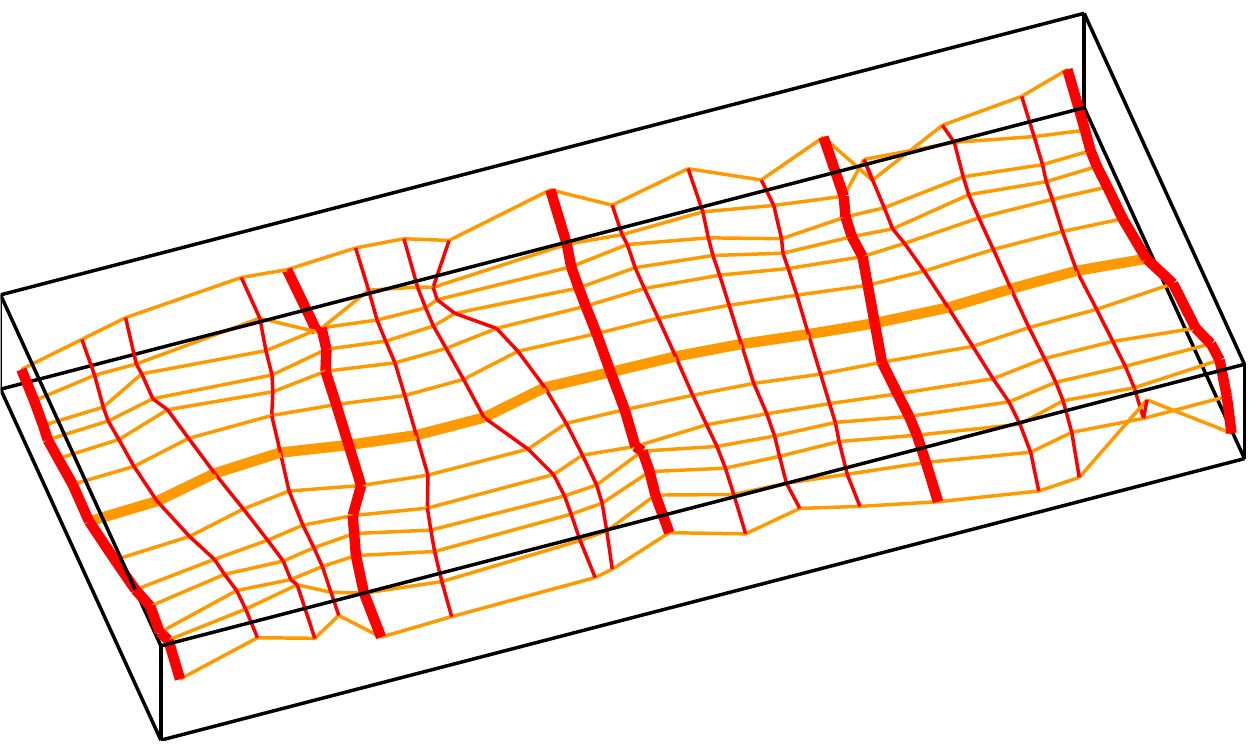} \\
\hspace{-1.5cm}\begin{turn}{90}
\begin{minipage}{3cm}
\begin{center}\small{Identified Features}\end{center}
\end{minipage}
\end{turn} \hspace{-1.5cm}
       & \includegraphics[width=3cm]{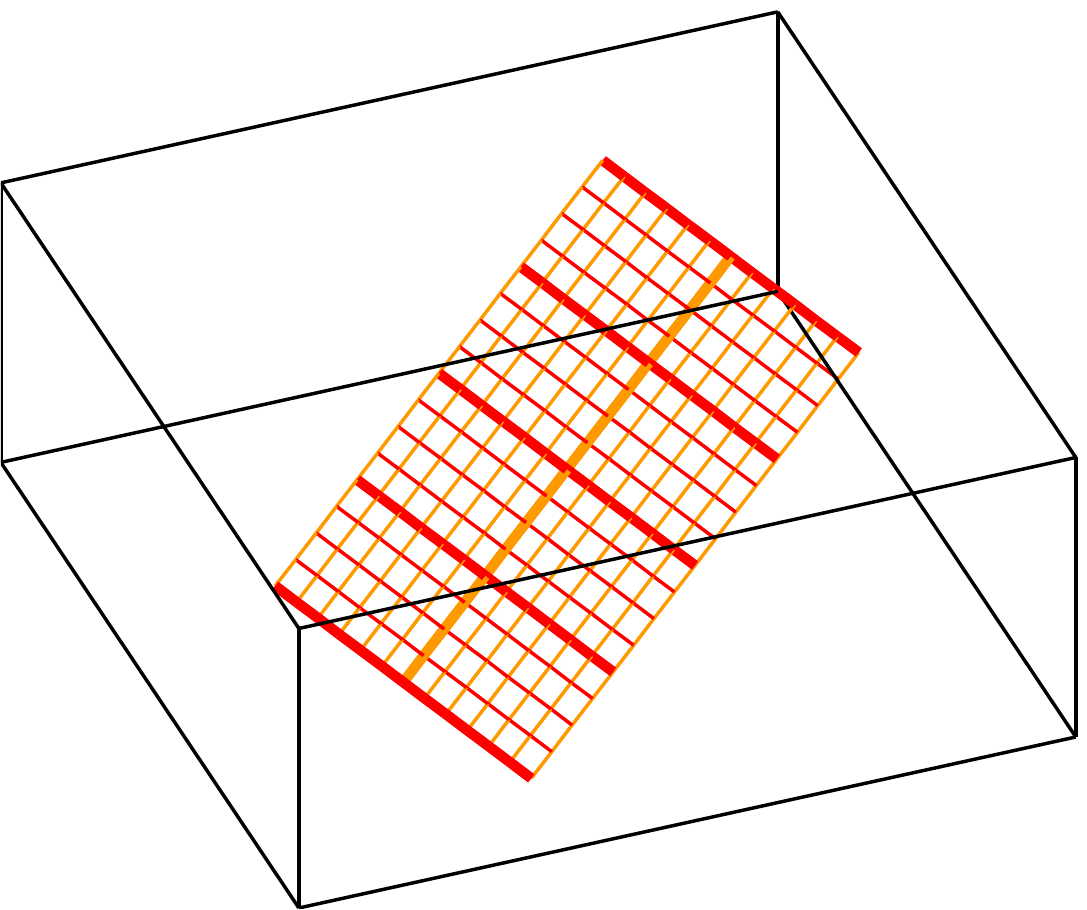} &
       \includegraphics[width=3cm]{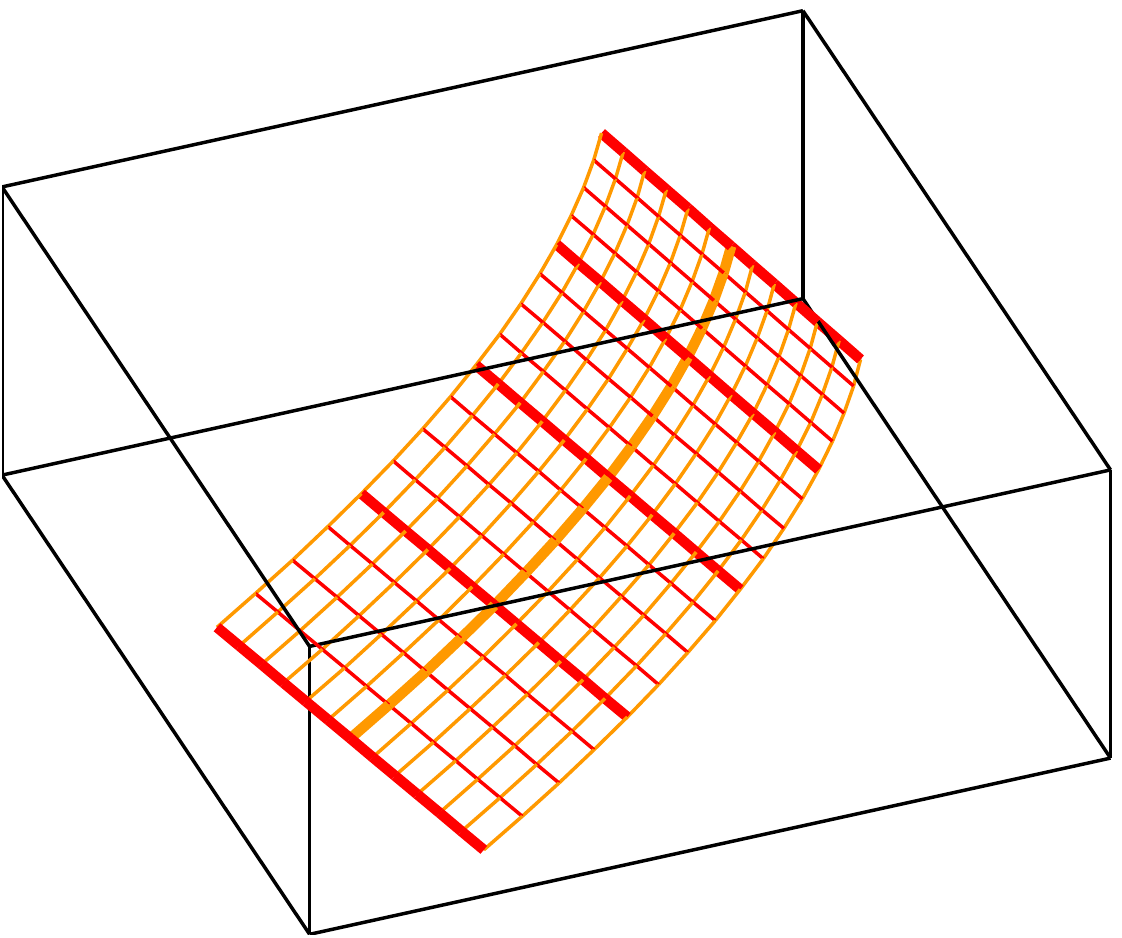} &
        \includegraphics[width=3cm]{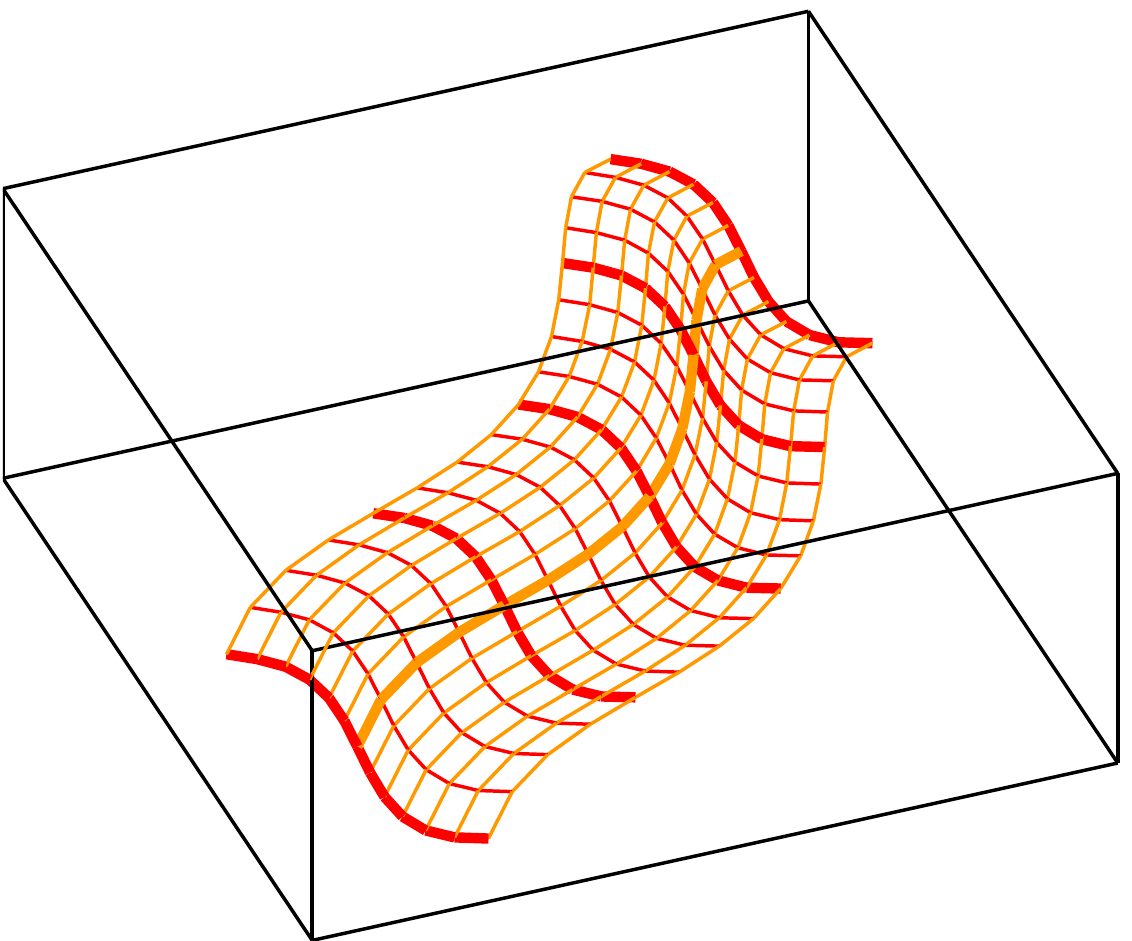} &
        \includegraphics[width=3cm]{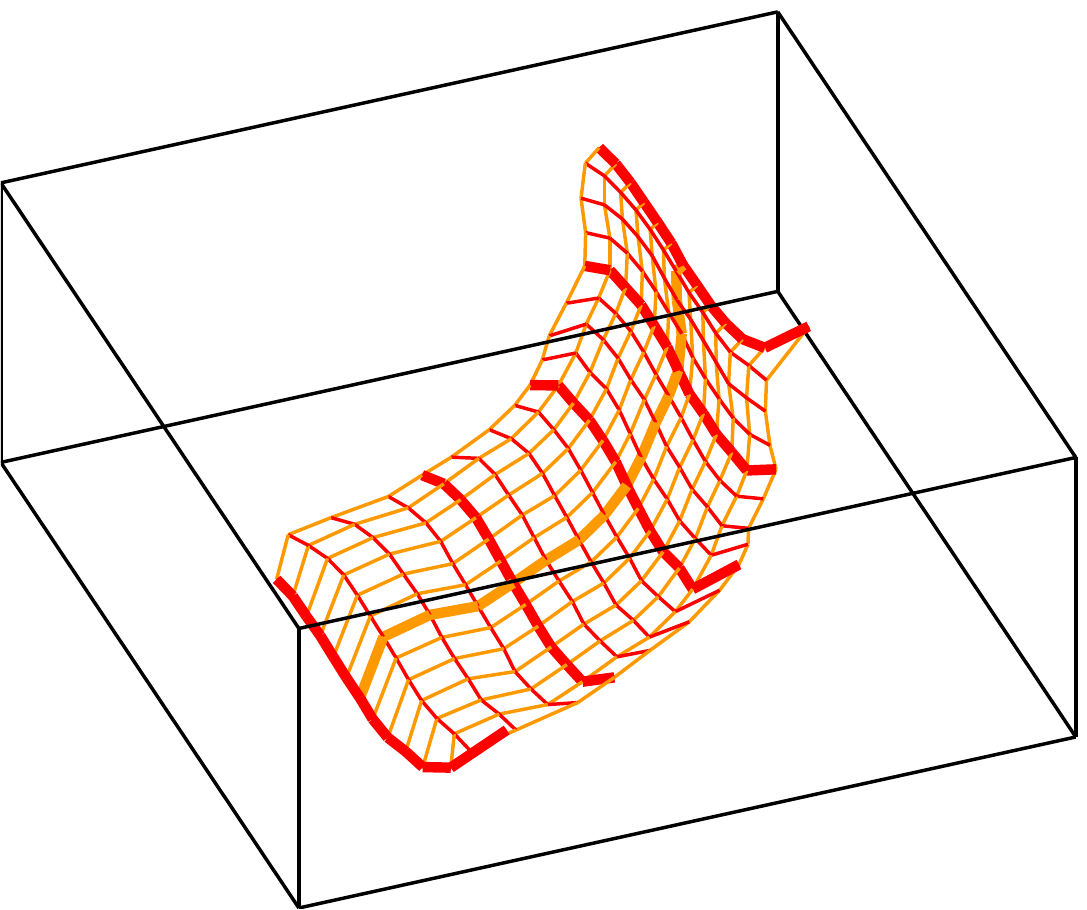} &
        \includegraphics[width=3cm]{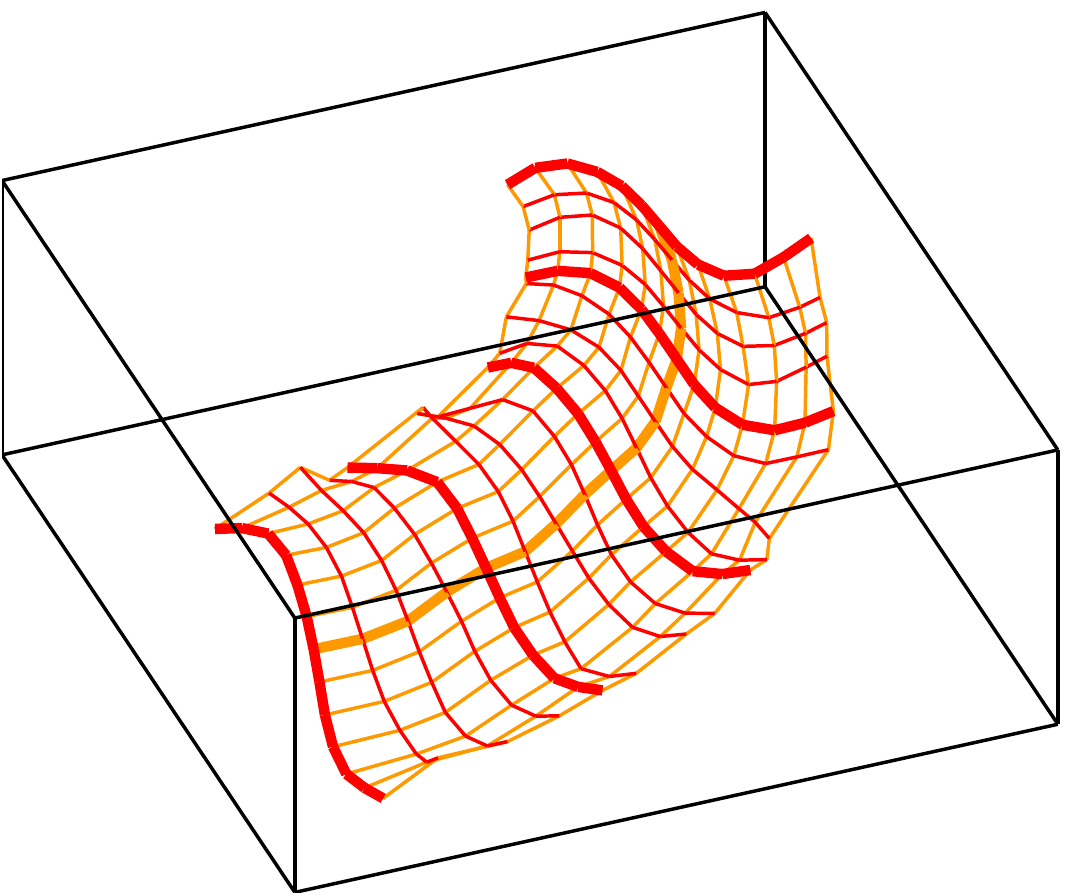} \\ [5mm]
    \hspace{-1.5cm}  &  \small \hspace{-1.9cm} {\emph{MSE}$_{{\bf DR}}$} \hspace{0.7cm} 100 & \small 48 $\pm$ 2 &\small 23.1 $\pm$ 0.6 &\small 3.3 $\pm$ 0.3 & \small 17 $\pm$ 3\\
    \hspace{-1.5cm}  &  \small \hspace{-1.7cm} \emph{MSE}$_{{\bf F}}$ \hspace{0.7cm} 100 & \small 66.4 $\pm$ 0.7 &\small 42.1 $\pm$ 0.4 &\small 49 $\pm$ 1 & \small 12 $\pm$ 1\\
	\\ [2mm]
        \hline
	\\ [2mm]
    \hspace{-1.5cm} & \small PCA  & \small NLPCA  &\small PPA  &\small DRR & \small SPCA    \\[-9mm]
\hspace{-1.5cm}\begin{turn}{90}
\begin{minipage}{3cm}
\begin{center}\hspace{-1.5cm} \small{Transform}\end{center}
\end{minipage}
\end{turn} \hspace{-1.5cm}
     & \includegraphics[width=3cm]{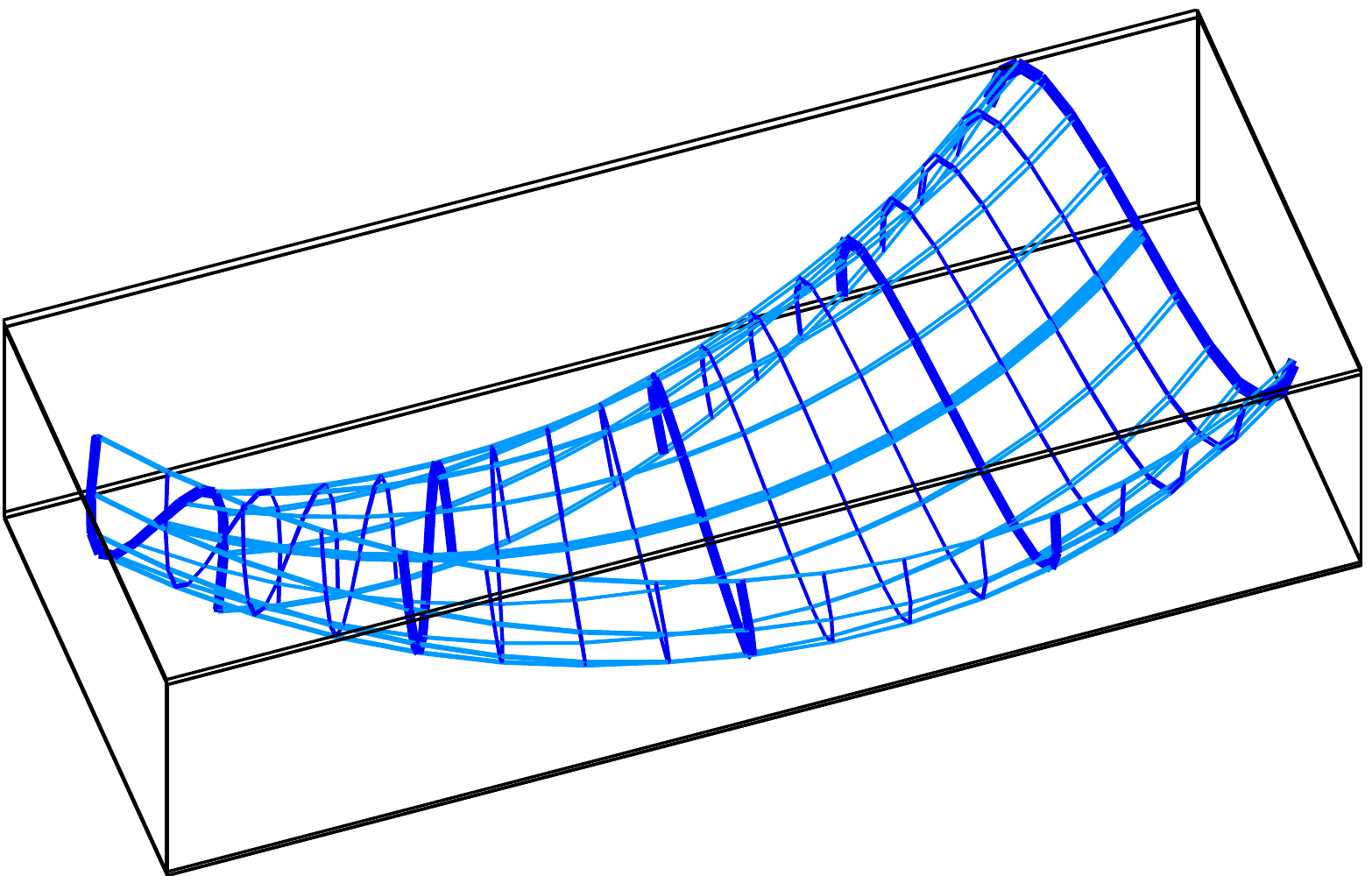} & 
       \includegraphics[width=3cm]{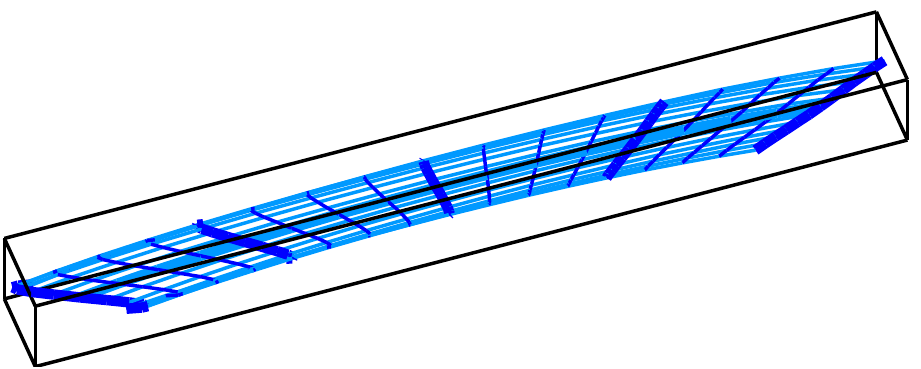} &
        \includegraphics[width=3cm]{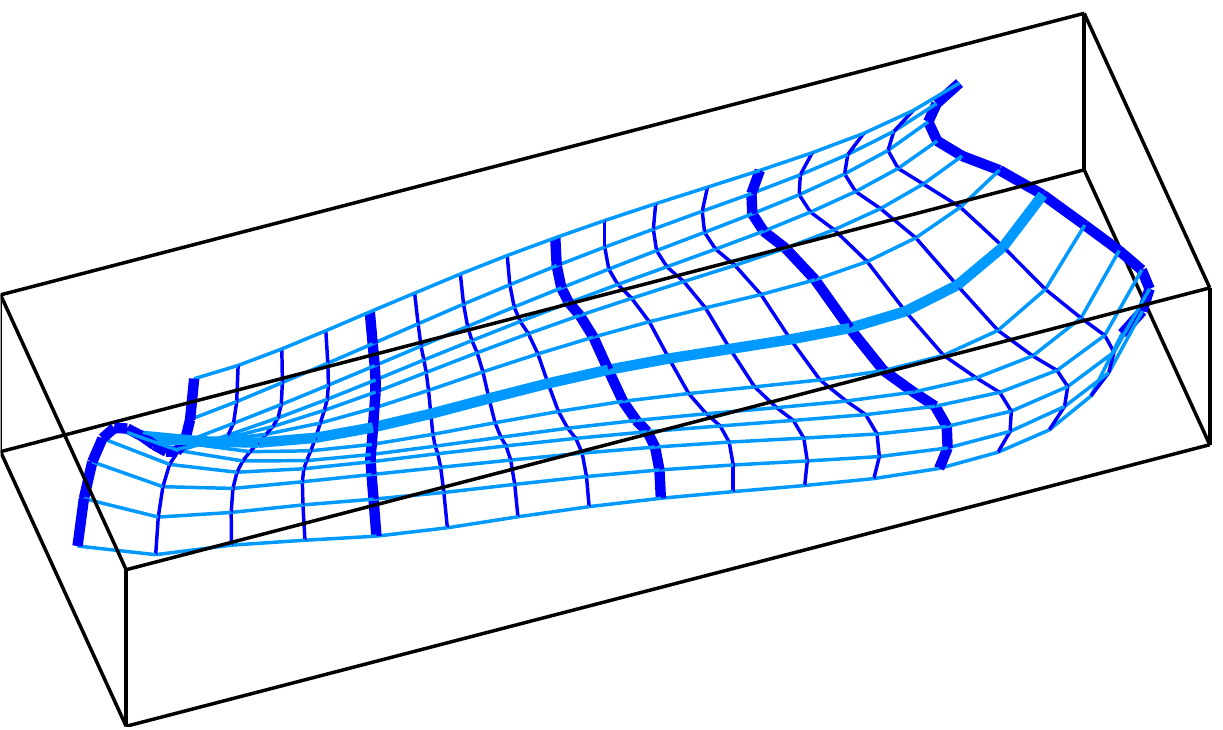} &
        \includegraphics[width=3cm]{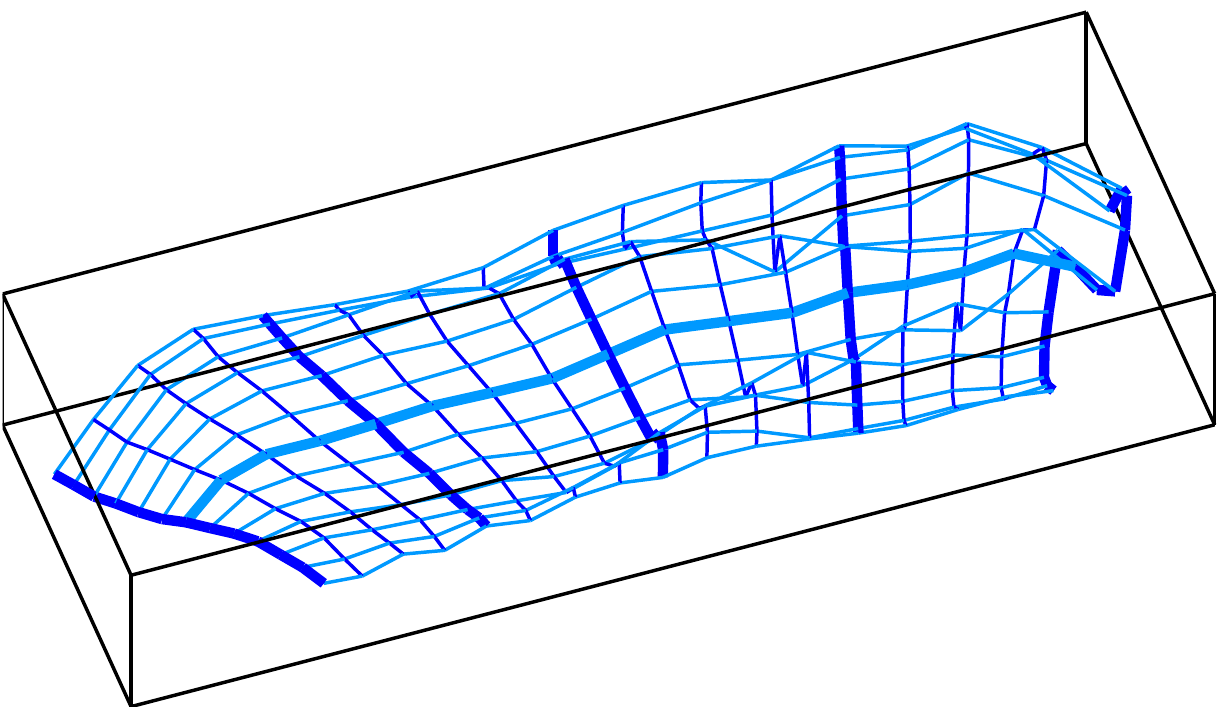} &
        \includegraphics[width=3cm]{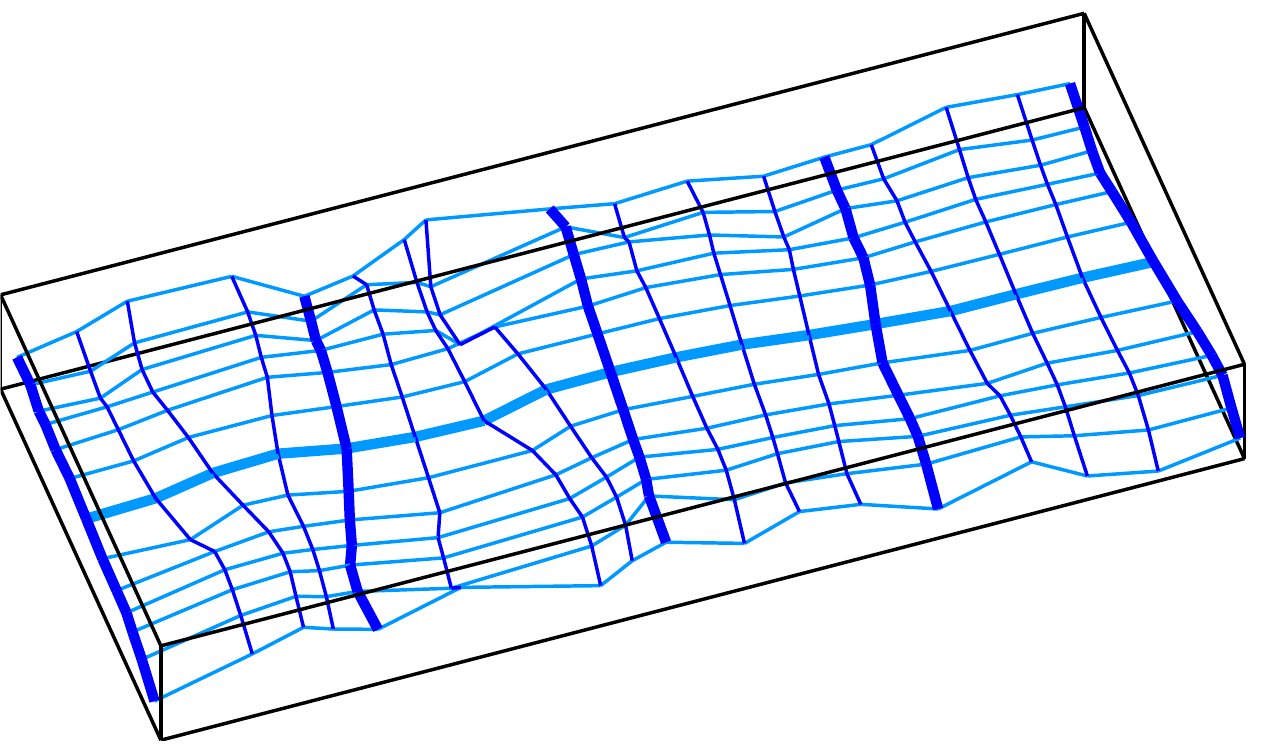} \\
\hspace{-1.5cm} \begin{turn}{90}
\begin{minipage}{3cm}
\begin{center}\small{Identified Features}\end{center}
\end{minipage}
\end{turn} \hspace{-1.5cm}
       & \includegraphics[width=3cm]{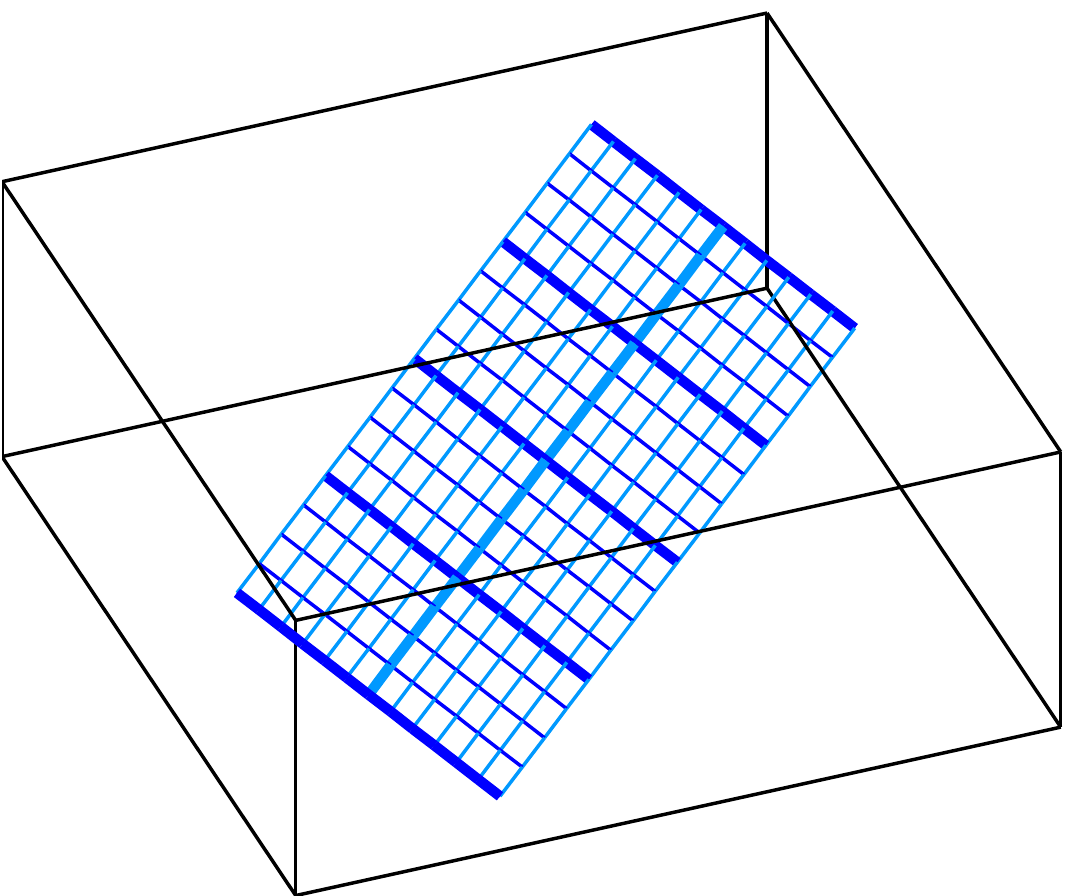} &
       \includegraphics[width=3cm]{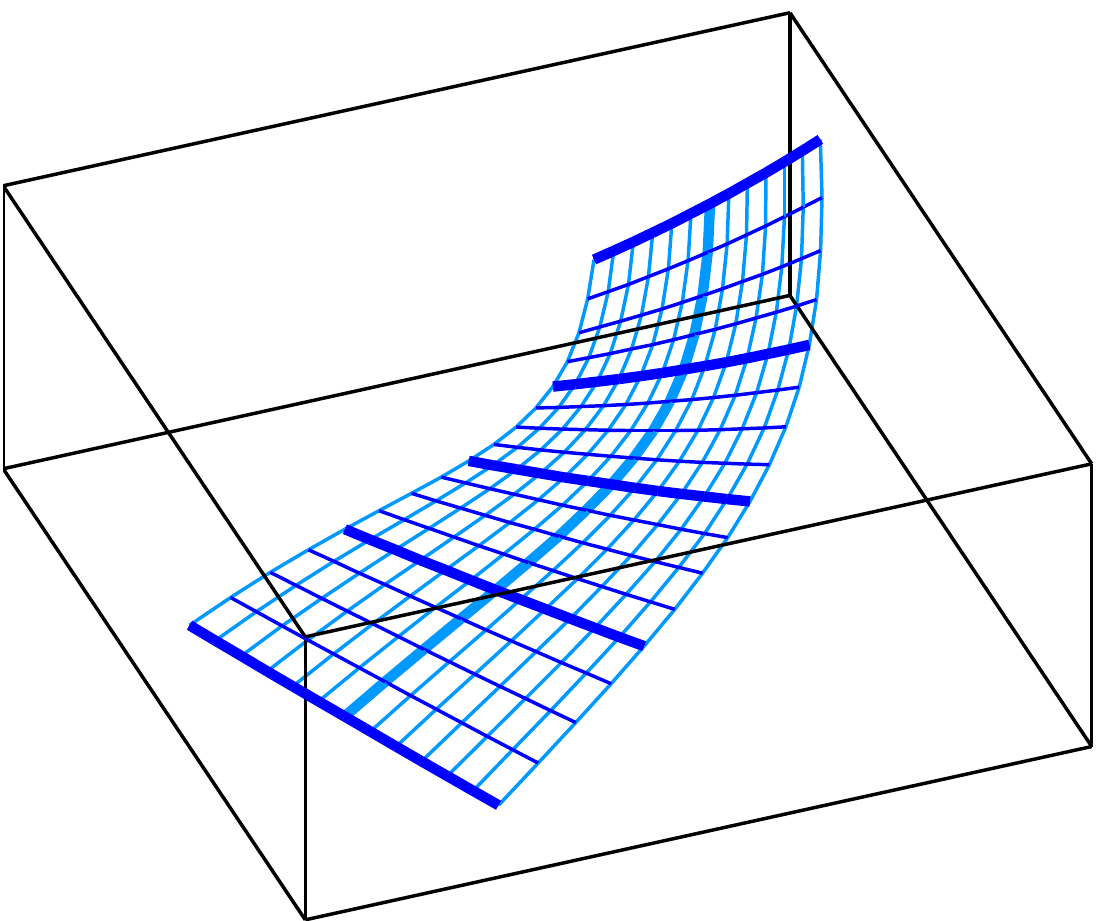} &
        \includegraphics[width=3cm]{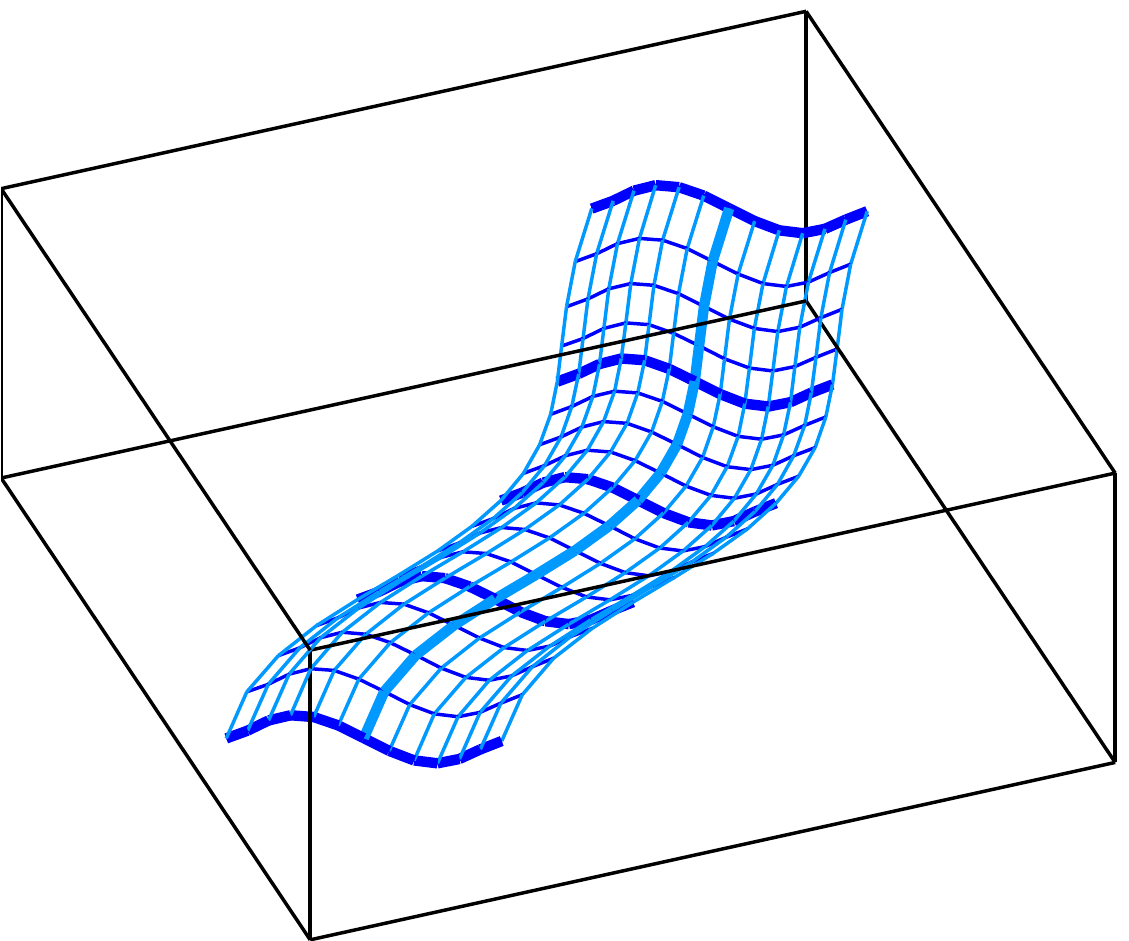} &
        \includegraphics[width=3cm]{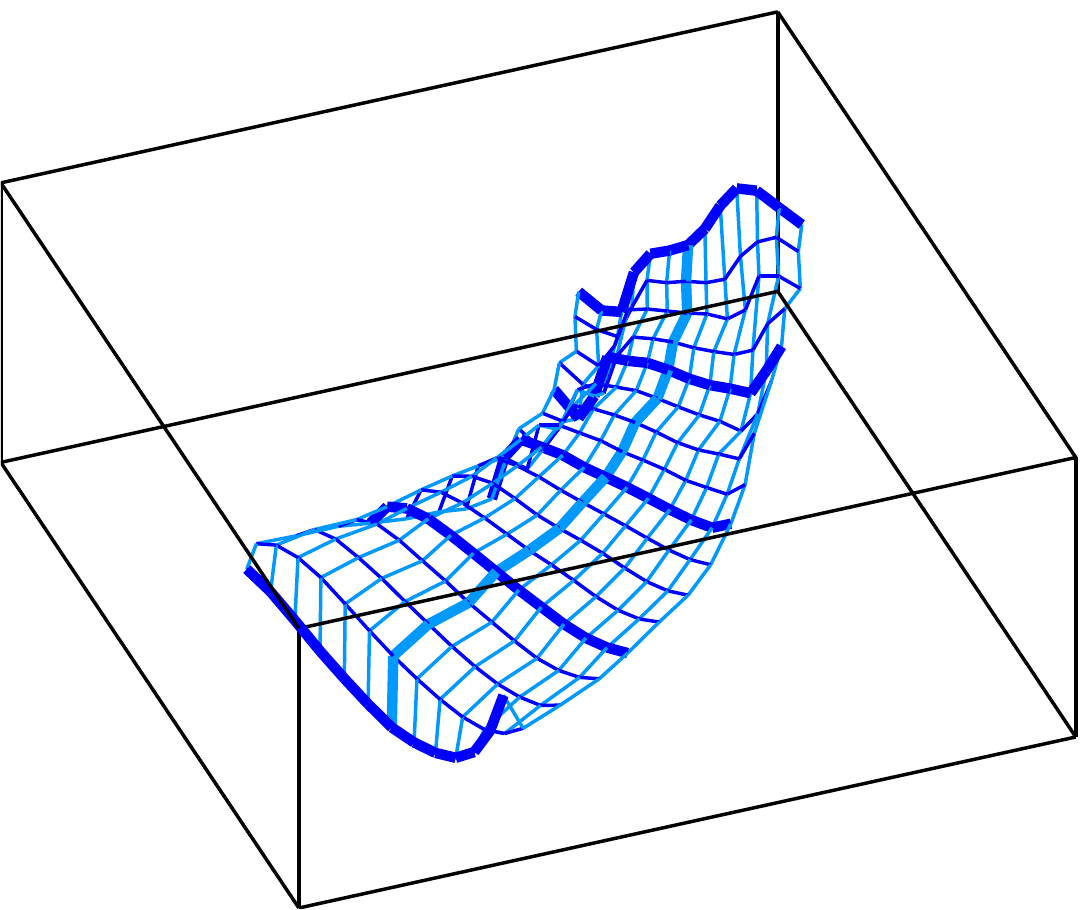} &
        \includegraphics[width=3cm]{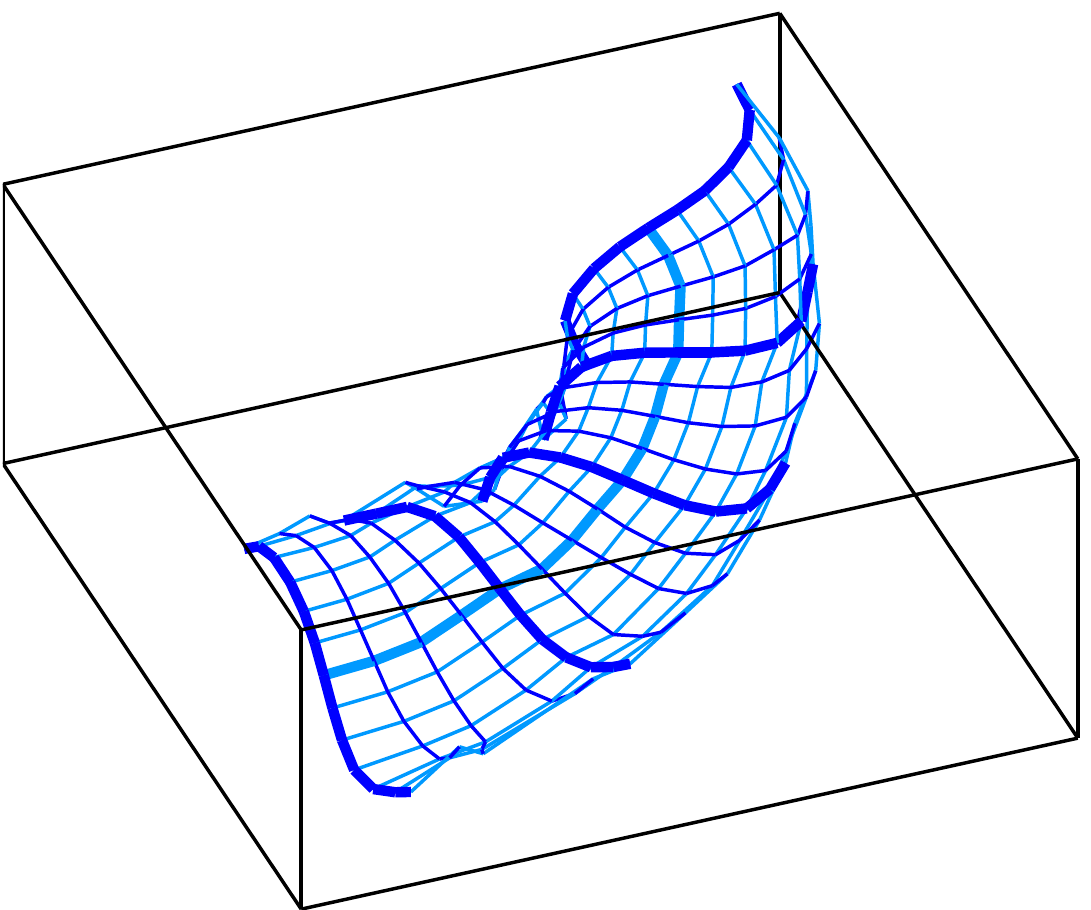} \\ [5mm]
    \hspace{-1.5cm} & \small \hspace{-1.6cm} {\emph{MSE}$_{{\bf DR}}$} \hspace{0.5cm} 269 $\pm$ 4 & \small  80 $\pm$ 40 &\small 97 $\pm$ 2 &\small  26.5 $\pm$ 0.3 & \small  19 $\pm$ 3\\
    \hspace{-1.5cm} & \small \hspace{-1.4cm} \emph{MSE}$_{{\bf F}}$ \hspace{0.5cm} 209 $\pm$ 2 & \small  85 $\pm$ 30 &\small 85.6 $\pm$ 0.4 &\small 69 $\pm$ 2 & \small 15 $\pm$ 4\\
    \end{tabular}
	\caption{\small Performance of the family of invertible representations on illustrative manifolds of different complexity. Complexity of the considered manifolds (top panel) depends on the tilt in secondary principal curves along the first principal curve~\cite{Delicado01}. Sample data are shown together with the first and secondary principal curves generated by the latent variables (angle and radius) in the input domain.
    Results of the different techniques for the considered manifolds are depicted in the same color as the input data (red for the no-tilt manifold, and blue for the tilted manifold). Previously reported representations range from rigid schemes such as PCA~\cite{Jolliffe02} to flexible schemes such as SPCA~\cite{Laparra12a,Laparra14b}, including practical nonlinear generalizations of PCA such as NLPCA~\cite{Scholz07} and PPA~\cite{Laparra14a} which are examples of intermediate flexibility between the extreme cases.
    Performance is compared in terms of reconstruction error when removing the third dimension (dimensionality reduction {\emph{MSE}$_{{\bf DR}}$} numbers are relative to the PCA error in the easy case), and in terms of the mean squared distance between the identified and the actual curvilinear features ({\emph{MSE}$_{{\bf F}}$} numbers are relative to the PCA error in the easy case).
    {\emph{MSE}$_{{\bf DR}}$} is related to the unfolding ability of the model (see the \emph{Transform} rows), and {\emph{MSE}$_{{\bf F}}$} is related to its ability to get appropriate curvilinear features from an assumed latent cartesian grid (see the \emph{Identified Features} rows).
    We used $10^4$ training samples and optimal settings in all methods.
    Dimensionality reduction was tested on the $17\times13$ highlighted curvilinear grid sampled from the true latent variables.
    The features in the input space were identified by inverting a $17\times13$ 2-$d$ cartesian grid in the transform domain.
    This (assumed) latent grid was scaled in every representation to minimize \emph{MSE}$_{{\bf F}}$.
    Standard deviations in errors come from models trained on 10 different data set realizations.}
	\label{fig1}
\end{figure*}

The advantages of DRR are illustrated in Fig.~\ref{fig1} where
we compare representative invertible representations of this family on two curved
and noisy manifolds of the class introduced by Delicado~\cite{Delicado01} (in red and blue).
This class of manifolds, originally presented to illustrate the concept of
\emph{secondary principal curves}~\cite{Delicado01}, is convenient since one can easily
control the complexity of the problem by introducing tilt (non-stationarity)
on the secondary principal curves (dark color) along the first principal curve (light color).
This controlled complexity is useful to point out the limitations of previous techniques
(e.g. required symmetry in the manifold) and how these limitations are alleviated by
using the (more general) DRR model.
The performance is compared in the input domain through the dimensionality reduction error and through the accuracy of the identified curvilinear features.
These manifolds come from a two-dimensional space of latent variables (positions along the first and secondary curves).
As a result, the dimensionality reduction error depends on the unfolding ability of the forward transform:
the closer the transformed data fit a flat rectangle, the smaller the error when truncating the representation.
On the other hand, the identified features depend on how the inverse transform bends a cartesian grid in the latent space: the better the model represents the curvature of data, the bigger the fidelity of the identified features.

Let us start by considering the performance on the easy case: manifold in red with no tilt along the second principal curve.
The previously reported techniques perform as expected:
on the one hand, progressively more flexible techniques (from PCA to SPCA) reduce the distortion after dimensionality reduction
(in \emph{MSE}$_{DR}$ terms) because they better unfold test data.
As a result, removing the third dimension in the rigid-to-flexible family progressively introduces less error.
On the other hand, the identified features in the input domain are progressively more similar to
the actual curvilinear latent variables when going from the rigid to the flexible extremes. 
In this specific \emph{easy} example the proposed DRR outperforms even the flexible SPCA in \emph{MSE}$_{DR}$ terms.
Moreover, since this particular manifold may not require increased flexibility (and hence may be better suited to the PPA model), 
PPA slightly outperforms DRR in \emph{MSE}$_{F}$ terms. 

Results for the more complex manifold (tilted secondary curves, in blue) provide more insight into the techniques since it pushes them (specifically PPA) to their limits.
Firstly, note that the increase in complexity is illustrated by an increase in the errors in all methods.
For instance, linear PCA, that identifies the same features in both cases, doubles the normalized MSEs.
While the reduction in performance is not that relevant in SPCA (remember these flexible techniques cannot be applied in high dimensional scenarios), this twisted manifold certainly challenges fast generalizations of PCA:
the MSEs dramatically increase for NLPCA and PPA.
Even though NLPCA identifies certain tilt in the secondary feature along the first curve, it seems too rigid to follow the data structure.
PPA displays a different problem: as stated above, by construction, the $i$-th curvilinear feature in PPA cannot handle relations with the $(i-1)$-th curve beyond the prediction of the mean.
This is because the data in all orthogonal subspaces along the $(i-1)$-th curve collapse, and
are described by a single curve depending on a single parameter (\emph{univariate regression}).
This leads to using the same $i$-th curve all along the $(i-1)$-th feature
(note the repeated secondary curves along the first curve in both, red and blue, cases).
This is good enough when data manifolds have the required symmetry (PPA performance is over NLPCA in the first case), but leads to dramatic errors when the method have to deal with relations between three or more variables, as for the manifold in blue, where PPA performance is below NLPCA. This latter effect frequently appears in hyperspectral images, as different (non-stationary) nonlinear relations between spectral channels occur for different objects~\cite{bachmann05,bachmann06,Camps11}.

Finally, note that DRR clearly improves PPA in the challenging example in blue, mainly because it uses multiple dimensions (instead of a single one) to predict each lower variance dimension in the data.
As a result, it can handle non-stationarity along the principal curves leading to better unfolding
(lower \emph{MSE}$_{DR}$) and tilted secondary features (lower \emph{MSE}$_{F}$). This removes the symmetry requirement in PPA and broadens the class of manifolds suited to DRR.

\section{Dimensionality Reduction via Regression}
\label{the_math_details}

PCA removes the second order dependencies between the signal components, i.e. PCA scores are decorrelated~\cite{Jolliffe02}.
Equivalently, PCA can be casted as the \emph{linear} transform that minimizes reconstruction error
when a fixed number of features are neglected.
However, for general non-Gaussian sources, and in particular for hyperspectral images,
PCA scores still display significant statistical relations, see~\cite{Camps11}[ch. 2].
The scheme presented here tries to \emph{nonlinearly} remove the information still shared by different PCA components.

\vspace{-0.0cm}
\subsection{DRR scheme}
It is well known that, even though PCA leads to a domain with decorrelated dimensions,
complete independence (or non redundant coefficients) is guaranteed only if the signal has a Gaussian probability density function (PDF).
Here, we propose a scheme to remove this redundancy (or uninformative data).
The idea is simple: just predict the redundant information in each coefficient that can be extracted from the others. Only the non-predictable information (the residual prediction error) should be retained for data representation.
Specifically, we start from the linear PCA representation outlined above: $\alpha = \V x$.
Then, we propose to predict each coefficient, $\alpha_i$, through a
multivariate regression function, $f_i(\cdot)$, that takes the higher variance components 
as inputs for prediction. The non-predictable information is:

\begin{equation}
     y_i = \alpha_i - \hat{\alpha}_i = \alpha_i - f_i(\alpha_{1},\alpha_{2},...,\alpha_{i-1}),
     \label{ith_step}
\end{equation}
and this residual, $y_i$, is the $i$-th dimension of the DRR domain.
This \emph{prediction+substraction} is applied $d-1$ times, $\forall \, i = d, \, d-1, \ldots, 2$, where $d$ is the dimension of the input.
As a result, the DRR representation of each input vector $x$, is:
\vspace{-0.0cm}
$$r = {\bf R}(x) = (\alpha_1, y_2, y_3, \ldots, y_d)^\top.$$

\vspace{-0.2cm}
\subsection{Properties of DRR}

\paragraph{DRR generalizes PCA}

In the particular case of using linear regressions in $f_i(\cdot)$, i.e. linear-DRR,
the prediction $\hat{\alpha}_i$ would be equal to zero. Note that this is the result when using classical (least squares) solution since ${\alpha}_i$ is uncorrelated with each ${\alpha}_1, {\alpha}_2, \ldots, {\alpha}_{i-1}$. Therefore $f_i({\alpha}_1, {\alpha}_2, \ldots, {\alpha}_{i-1})=0$, and then $y_i = \alpha_i$, i.e. linear-DRR reduces to PCA.

As a result, if the employed nonlinear functions $f_i(\cdot)$ generalize the linear functions,
DRR will obtain at least as good results as PCA.
The flexibility of these functions with regard to the linear case will reduce the variance
of the residuals, and hence the reconstruction error in dimensionality reduction.

\paragraph{DRR is invertible}

Given the DRR transformed vector, $(\alpha_1, y_2, y_3, \ldots, y_d)^\top$, and knowing the functions of model $f_i(\cdot)$,
the inverse is straightforward since it reduces to sequentially undo the forward transformation:
we first predict coefficient $(i+1)$-th from previous (known) coefficients using the known regression function,
and then, we use the known residual to correct the prediction:
\begin{equation}
      \alpha_i = \hat{\alpha}_i + y_i = f_i(\alpha_{1},\alpha_{2},...,\alpha_{i-1}) + y_i
      \label{inverse}
\end{equation}

\paragraph{DRR has an easy out-of-sample extension}

Note that forward and inverse DRR transforms can be applied to new data (not in the training set)
since there is no restriction to apply the prediction functions $f_i(\cdot)$. See Sec. \ref{KRR} for a discussion on the selected regression functions in this work.

\paragraph{DRR is a volume preserving transform}

A nonlinear transform preserves the volume of the input space if the determinant of its Jacobian is one for all $x$ \cite{Dubrovin82}.
Here we prove that the nature of DRR ensures that its Jacobian fulfills this property.

DDR can be thought of as a sequential algorithm in which only one dimension is addressed at a time
through the elementary transform ${\bf R}_i$ consisting of prediction and substraction (Eq. \eqref{ith_step}).
Yet mathematically convenient to formulate the Jacobian, this sequential view is does not prevent the parallelization discussed later.
Hence, the (global) Jacobian of DRR, $\nabla {\bf R}$, is the product of the Jacobians of the elementary transforms in this sequence times the matrix of the initial PCA as follows:
$$
  \nabla {\bf R}(x) = \left(\prod_{i=2}^{d} \nabla {\bf R}_i \right) \V.
$$

The $i$-th elementary transform leaves all, but the $i$-th dimension, unaltered. Therefore, each elementary Jacobian is the identity matrix except for the $i$-th row, which depends on
$\alpha_1, \cdots, \alpha_{i-1}$ through the derivatives of the $i$-th regression function with regard to each component in the previous stage:
$$
\nabla {\bf R}_i = \left(
  \begin{array}{ccccccccc}
    1 &   &        &   &   &   &   &   &   \\
      & 1 &        &   &   &   &   &   &   \\
    0 &   & \ddots &   &   &   & 0 &   &   \\
      &   &        & 1 &   &   &   &   &   \\
    \frac{-\partial f_i}{\partial \alpha_1} & \frac{-\partial f_i}{\partial \alpha_2} & \cdots & \!\!\!\!\frac{-\partial f_i}{\partial \alpha_{i-1}} & 1 &  &  &  &  \\
      &   &        &   &   & 1 &   &   &  \\
      &   &   0    &   &   &   & \ddots &  &  \\
      &   &        &   &   &   &   & 1 &   \\
      &   &        &   &   &   &   &   & 1 \\
  \end{array}
\right)
$$

Whatever these derivatives are (whatever regression function is selected), the determinant of such a simple matrix is always one at every point $x$. Therefore, the determinant of the global Jacobian is guaranteed to be one including the PCA transform, $\V$, which is orthonormal.

\paragraph{Parallelization of DRR}

Multivariate regression in DRR is a qualitative advantage over other methods (as discussed in Section \ref{related}). However it is computationally expensive. Fortunately, the proposed DRR allows trivial parallel implementation of the forward transform.
Note that the prediction of each component is actually done from a subset of the original PC scores.
Therefore, all the prediction functions, $f_i(\cdot)$, can be applied at the same time after the initial PCA step, and sequential implementation is not necessary.
This is an obvious \emph{computational advantage} over PPA, which necessarily requires a sequential implementation,
but it represents a \emph{qualitative advantage} too, since in PPA each feature depends on the previous nonlinear features (see Eq. \ref{PPAapprox}), while in DRR nonlinear regressions
only depend on linear features, but not on previous curvilinear coefficients.
As opposed to the forward transform, the inverse is not parallelizable since,
in order to predict the $i$-th PCA coefficient, we need the previous linear PCs, which implies operating sequentially from $i = 2, \ldots, d$.

\subsection{Selecting the class of regression functions}
\label{KRR}
In practice, the prediction functions $f_i(\cdot)=\hat{\alpha_i}$ reduce to training a set of nonlinear regression models. In our experiments, we used the kernel ridge regression (KRR)~\cite{shawetaylor04} to implement the prediction functions $f_i(\cdot)$, although any alternative regression method could be also applied. Notationally, given $N$ data points, the prediction for all of them is estimated as:
$$
\hat{\alpha_i} = \sum_{j=1}^N{k(\alpha_{\backslash i},\A_{j,\backslash i})}\,\,\,\beta_{j},
$$
where $k(\cdot,\cdot)$ is a kernel (similarity) function reproducing a dot product in Hilbert space, $\alpha_{\backslash i} = (\alpha_1,\ldots, \alpha_{i-1})$, $\A\in\Real^{N \times d}$ is the matrix containing all the $N$ training samples in rows, $\A_{i}\in\Real^{N\times 1}$ is the $i$-th column of $\A$ to be estimated, $\A_{\backslash i}\in\Real^{N\times (i-1)}$ denotes a submatrix containing columns $1, \ldots, i-1$ of $\A$ used as input data to fit the prediction model, and $\A_{j,\backslash i}\in\Real^{1\times(i-1)}$ represents the feature vector in row $j$ of $\A_{\backslash i}$. In this prediction function, $\boldsymbol{\beta}\in\Real^{N\times 1}$ is the dual weight vector obtained by solving the least squares linear regression problem in Hilbert space:
$$\boldsymbol{\beta} = (\K+\gamma {\bf I})^{-1}\A_{i},$$
where $\K\in\Real^{N\times N}$ is the kernel matrix with entries $\K_{lm}=k(\A_{l,\backslash i},\A_{m,\backslash i})$, being $l,m=1,\ldots,N$. Two parameters must be tuned for the regression: the regularization parameter $\gamma$ and the kernel parameters. In our experiments we used the standard squared exponential kernel function, $k(a,b)=\exp(-\frac{1}{2\sigma^2}\|a-b\|^2)$, as it is a universal kernel which involves estimating only the length-scale $\sigma$. Both $\sigma$ and $\gamma$ can be estimated by standard cross-validation.

KRR can be quite convenient in the DRR scheme because it implements flexible nonlinear regression functions, and reduces to solving a matrix inversion problem with unique solution. KRR offers a moderate training and testing computational cost\footnote{While naive implementations scale as ${\mathcal O}(N^3)$ for training, recent sparse and low-rank approximations~\cite{Lazaro10sparse,Arenas13spm} along with divide-and-conquer schemes~\cite{Wainwright13} can make KRR very efficient.}, includes regularization in a natural way, and also offers the possibility to generate multi-output nonlinear regression. The latter is an important feature to extend the DRR scheme to multiple outputs approximation. 
Finally, KRR has been successfully used in many real applications~\cite{Suykens02,shawetaylor04} including remote sensing data analysis involving hyperspectral data~\cite{CampsValls10eumtgrs}. However, it should be noted that, even in such cases, a previous feature extraction was mandatory to attain significant results~\cite{CampsValls10eumspie,CampsValls10eumigarss,CampsValls10eumtgrs,Arenas13spm}.

\section{Experimental results}
\label{experiments}

In this section, we give experimental evidence of the performance of the proposed algorithm in two illustrative settings. First,
we show results on the truncation error in a multispectral image classification problem including spatial context. 
Then we evaluate the performance of DRR in terms of both the reconstruction error and the expressive power of the features to perform multi-output regression of a challenging problem involving hyperspectral infrared sounding data\footnote{Reproduction of the experiments in this work is possible using the generic DRR toolbox at, \url{http://isp.uv.es/drr.html}}.

Focusing in these two experiments is not arbitrary. The two applications imply challenging high dimensional data: (1) multispectral image classification in which contextual information is stacked to the spectral information highly increases the dimensionality, and (2) hyperspectral infrared sounding data used to estimate atmospheric state vectors is densely sampled.
In both cases the input space may become redundant because of the collinearity introduced either by the (locally stationary) spatial features or by the spectral continuity of natural sources. In these experiments, in which $d>35$, we compare DRR with members of the invertible projection family described in Section \ref{related} suited to high dimensional scenarios. This implies focusing on PCA, NLPCA and PPA, excluding SPCA and SOM because of their prohibitive cost.

\begin{figure*}[t!]
\small
\begin{center}
\begin{tabular}{cc}
\includegraphics[width = 8cm]{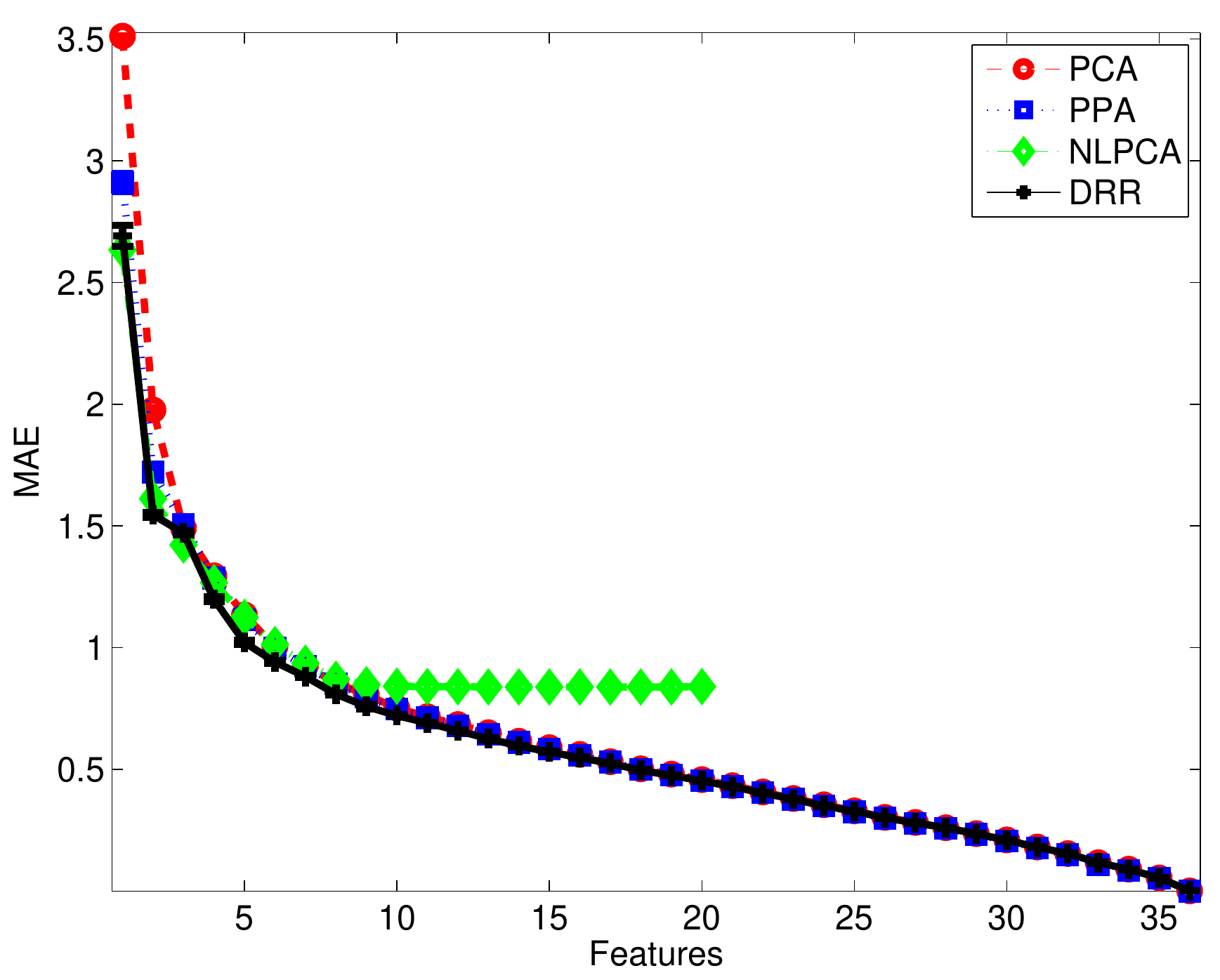} &
\includegraphics[width = 8cm]{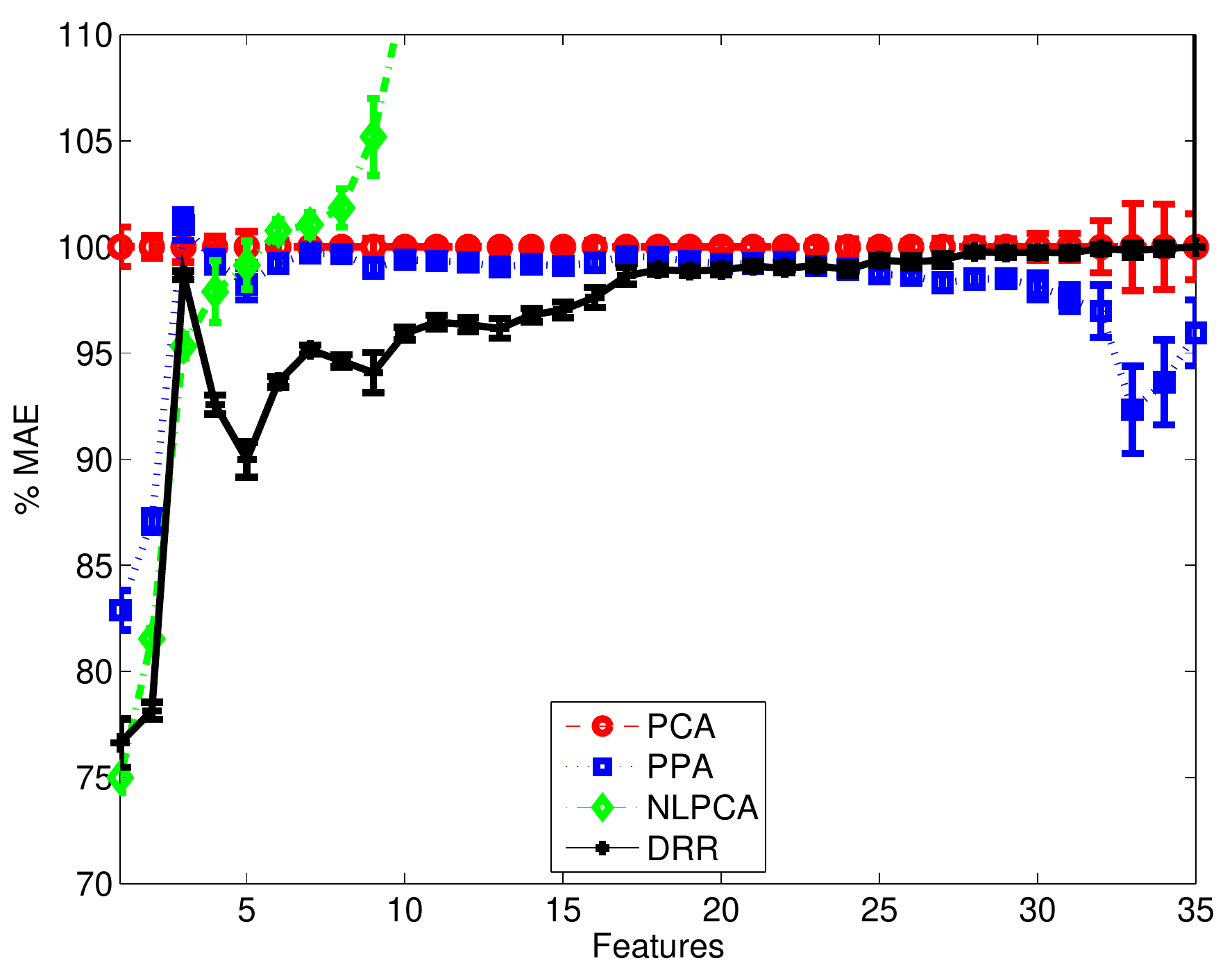}
\end{tabular}
\end{center}
\vspace{-0cm}
\caption{Reconstruction error results on the contextual multispectral image classification. Comparison between PCA, PPA, NLPCA and DRR for different number of extracted features, in both mean absolute reconstruction error (MAE) (left) and relative MAE with respect to PCA error (right), for which going below the PCA means better results (less error).}
\label{exp_DR}
\end{figure*}

\begin{figure*}[t!]
\small
\begin{center}
\begin{tabular}{cc}
\includegraphics[width = 8cm]{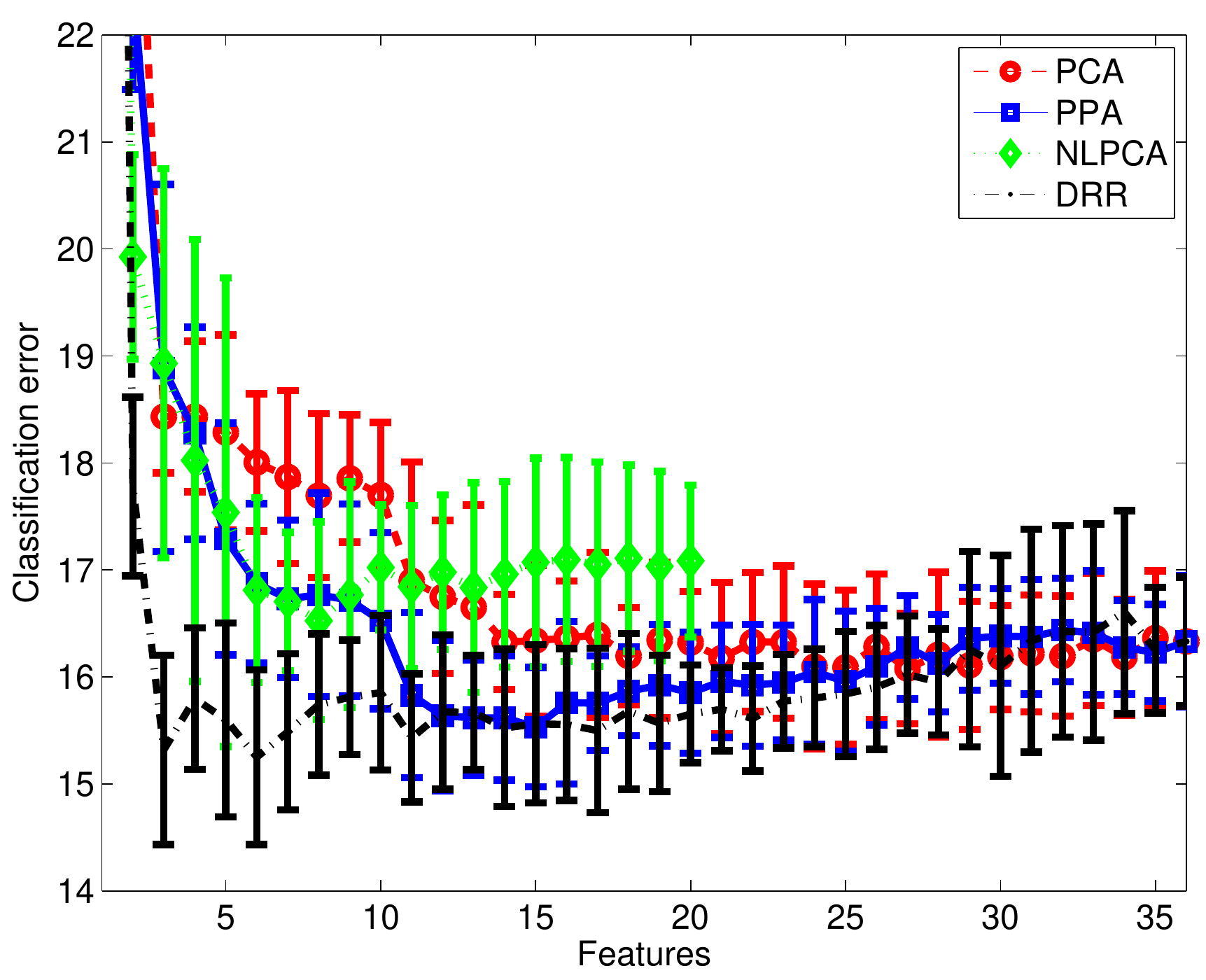} &
\includegraphics[width = 8cm]{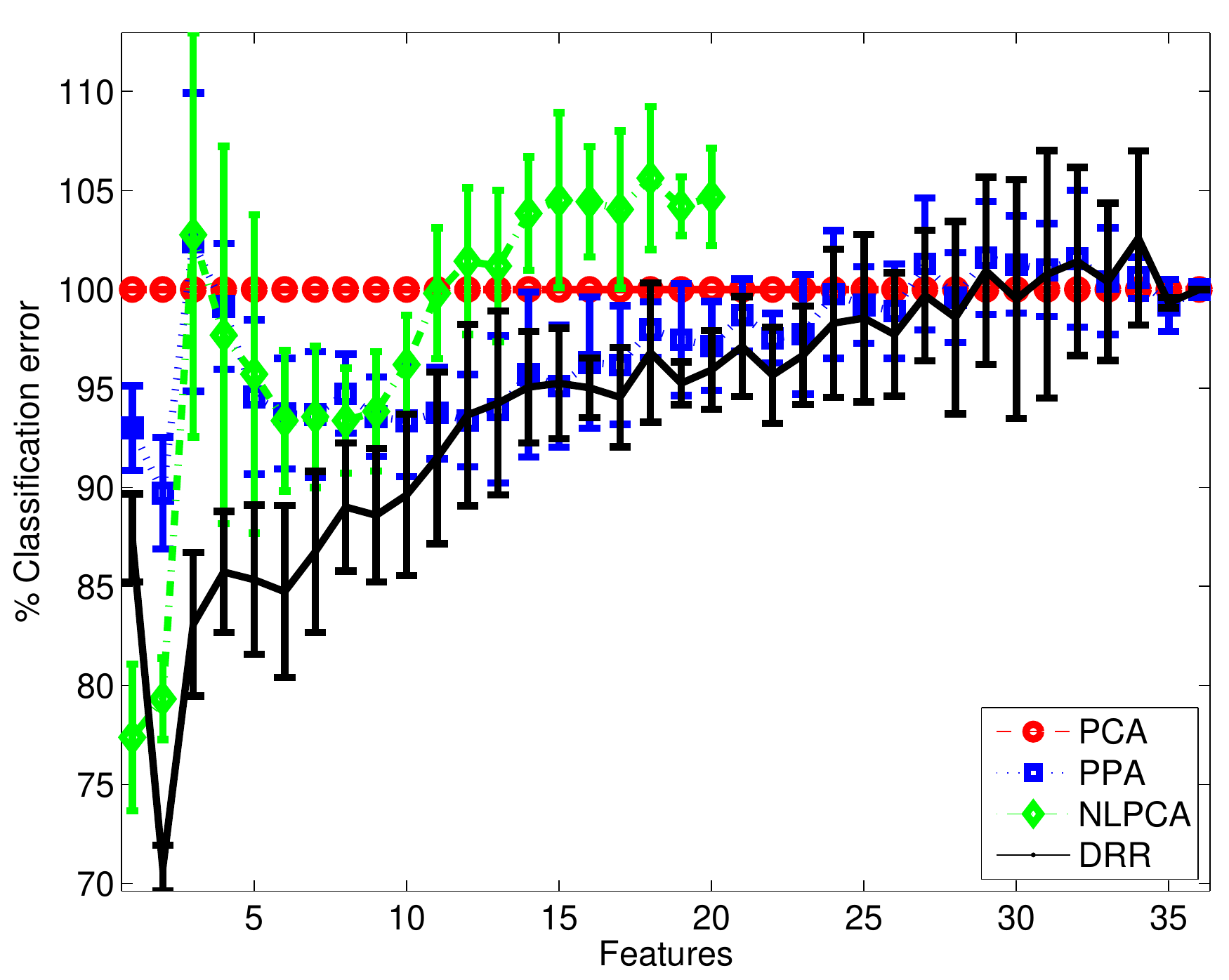}
\end{tabular}
\end{center}
\vspace{-0cm}
\caption{Classification results on the contextual multispectral image classification. Comparison between PCA, PPA, NLPCA and DRR for different number of extracted features, in both classification error (left) and relative classification error with respect to PCA accuracy (right), for which going below the PCA means better results (less error).}
\label{exp_DR2}
\end{figure*}

\subsection{Experiment 1: Multispectral image classification}

For our first set of experiments, we considered a Landsat MSS image consisting of $82\times 100$ pixels with a spatial resolution of 80m $\times$ 80m (all data acquired from a rectangular area approximately $8$ km wide) \footnote{Image available at http://www.ics.uci.edu/~mlearn/MLRepository.html}. Six classes are identified in the image, namely red soil, cotton crop, grey soil, damp grey soil, soil with vegetation stubble and very damp grey soil. A total of $6435$ labeled samples are available. Contextual information was included stacking neighboring pixels in 3$\times$3 windows. Therefore, $36$-dimensional input samples were generated, with a high degree of redundancy and collinearity. We address two problems with this dataset: a pure spatio-spectral dimensionality reduction problem, and the effect of the reduced dimension in image classification.

\subsubsection{Reconstruction accuracy}
In the first problem,
we compare the dimensionality reduction performance in terms of Mean Absolute Error (MAE) in the original domain.
Note that this kind of evaluation can be used only with invertible methods. For each method, the data are transformed and then inverted using less dimensions. This is equivalent to truncate dimensions in PCA. In order to illustrate the advantage of using a given method instead of PCA, results are shown in percentage with regard to the PCA performance:
$
	\% \text{MAE}_{\text{method}} = 100\,\,\text{MAE}_{\text{method}}/\text{MAE}_{\text{PCA}},
$
where $\text{MAE}_{\text{method}}$ and $\text{MAE}_{\text{PCA}}$ refer to the MAE obtained with the considered method and PCA, respectively.

Figure~\ref{exp_DR} shows the results of the experiment. We divided the available labeled data into two sets (training and test) with equal number of samples. The samples of each set have been randomly selected from the original image dataset. The MAE of reconstruction in the test set averaged over ten independent realizations is shown. Several conclusions can be obtained: Specifically, NLPCA obtains good results when a few number of extracted features are obtained, but rapidly degrades its performance with more than 10 extracted features, revealing a clear inability to handle high-dimensional problems. Note that the available implementation of NLPCA\footnote{http://www.nlpca.org/} is restricted to extract at most $20$ features. For a given number of extracted features, the reconstruction error increases substantially with regard to PCA (Fig.~\ref{exp_DR} right). PPA shows better results than NLPCA, and it is better suited than PCA in all the number of extracted features. Nevertheless, it is noticeable that DRR is in all cases better than all the other methods, revealing a maximum gain of +25\% over PCA for very few features.

\subsubsection{Classification accuracy}
The second problem with this dataset shows the classification results using the inverted data into the original input space of the different methods. We used the standard linear discriminant analysis on top of the inverted data\footnotetext{While other more sophisticated nonlinear classifiers could be used here, we are actually interested in this setting that allows us to study the expressive power of the extracted features. An homologous setting will be also used in the regression experiments of next subsection.}. In all cases, we used 3200 randomly selected examples for training and the same amount for testing. Test results are averaged over five realizations, and are shown in Fig.~\ref{exp_DR2}. The performance results indicate similar trends observed in the reconstruction error in Fig.~\ref{exp_DR}. Essentially, DRR outperforms the other methods, especially noticeable when a few number of components are used for reconstruction and classification. As the number of components increase, DRR and PPA show similar results. These results suggest that DRR better compacts the information in a lower number of components, which is useful for both reconstruction and data classification.
\subsubsection{Computational load}

Table \ref{time_SAT} shows the computation cost for all considered methods for training and testing\footnote{Experiments were performed using Matlab on an Intel 3.3 GHz processor with 48 GB RAM memory. No parallelization was applied on DRR in this experiment.}. The experiments used $3200$ training and $3200$ test samples, with $d=36$. Two main conclusions can be extracted: NLPCA is the most computationally costly algorithm for training and DRR for testing. 

\begin{table}[h]
\caption{Computational cost Landsat Dataset}
\begin{center}
\begin{tabular}{|l|l|l|l|l|}
\hline
              		& PCA 	& PPA 	& NLPCA	 & DRR \\ \hline \hline
Training time (sec) 	& 0.05 	&  0.6 	&   7944 & 1920 \\ \hline
Testing time (sec) 	& 0.007 &  0.16 &   0.05 & 35  \\ \hline
\end{tabular}
\label{time_SAT}
\end{center}
\end{table}

\subsection{Experiment 2: Regression from infrared sounding data}

We here analyze the benefits of using DRR for the estimation of atmospheric parameters from hyperspectral infrared sounding data with a reduced dimensionality.
We first motivate the problem, and then describe the considered dataset.
Again, we are interested in analyzing the impact of the reduced dimensionality both in the reconstruction error and in a different task, in this case, the retrieval of geophysical parameters.

Temperature and water vapor are atmospheric parameters of high importance for weather forecast and atmospheric chemistry studies \cite{Liou2002,Hilton2009}. Observations from spaceborne high spectral resolution infrared sounding instruments can be used to calculate the profiles of such atmospheric parameters with unprecedented accuracy and vertical resolution \cite{Huang1992}. In this work we focus on the data coming from the Infrared Atmospheric Sounding Interferometer (IASI), the Microwave Humidity Sensor (MHS) and the Advanced Microwave Sensor Unit (AMSU) onboard of the MetOp-A satellite\footnote{https://directory.eoportal.org/web/eoportal/satellite-missions/m/metop}. The IASI instrument is the one that poses the major dimensionality challenge due to its dense spectrum sampling: while MHS and AMSU spectra consist of about twenty values together, IASI spectra consist of $8461$ spectral channels, between $3.62$ and $15.5$~$\mu$m, with a spectral resolution of $0.5$~cm$^{-1}$ after apodization \cite{Chalon01,Simeoni1997}. Its spatial resolution is $25$~km at nadir with an Instantaneous Field of View (IFOV) size of $12$~km at an altitude of $819$~km.
This huge data dimensionality typically requires simple and computationally efficient processing techniques.

One of the retrieval techniques available in the MetOp-IASI Level 2 Product Processing Facility (L2 PPF) is a computationally inexpensive method based on linear regression from the principal components of the measured brightness spectra and the atmospheric state parameters. We aim to introduce DRR in such scheme as an alternative to PCA. In this application it is important that dimensionality reduction minimizes the reconstruction error and that the identified features are useful in the retrieval stage.

\begin{figure}[t!]
\centering
\setlength{\tabcolsep}{1pt}
\begin{tabular}{cc}
\includegraphics[width = 7cm]{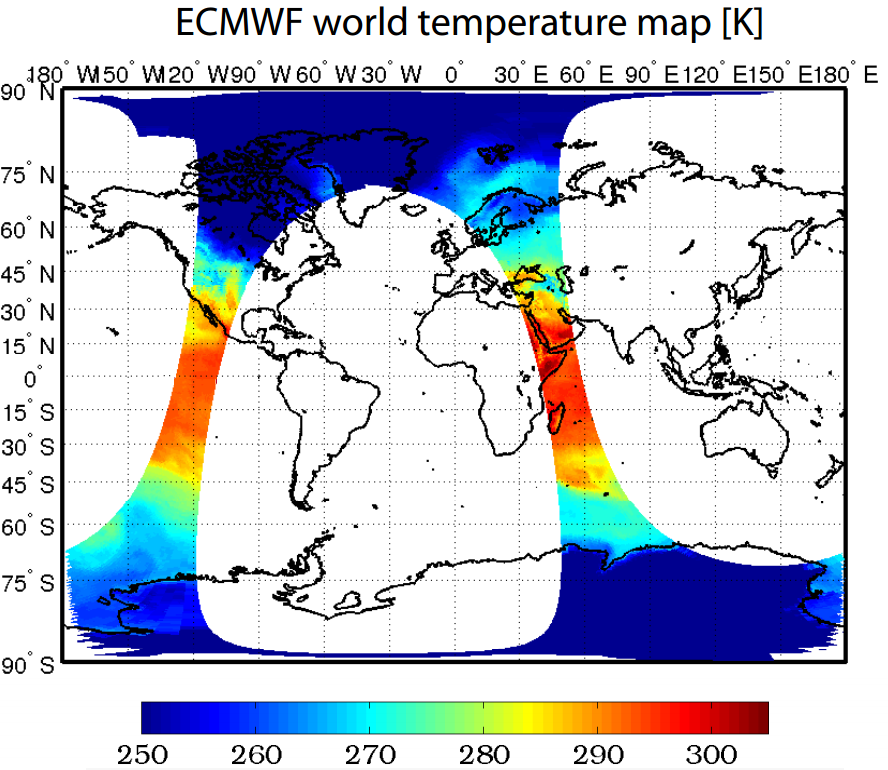}
\end{tabular}
\vspace{-0.25cm}
\caption{\label{fig:map} Surface temperature [in K] world map provided by the official ECMWF model, http://www.ecmwf.int/.
}
\end{figure}

We used a collection of $23$ datasets of input data from the different sensors: IASI, MHS and AMSU. The considered output atmospheric variables are diverse, e.g. temperature, moisture, and surface pressure.
In each dataset provided by EUMETSAT, the preprocessed input data were $110$-dimensional.
Each input vector consisted of the following: \emph{one} scalar indicating the secant of satellite zenith angle,
$19$ radiance values from the AMSU and MHS sensors, and $90$ values from the IASI sensor.
The data from IASI were actually three separate sets of $30$ PC scores each, from three different IASI bands.
Note that, despite intra-band decorrelation, the vector elements may still exhibit statistical dependency, which may be significant even at a second order level, among different bands and instruments.

The data to be predicted (or output data) is $277$-dimensional.
Each output vector consists of the following: $4$ data corresponding to the \emph{surface temperature}
and \emph{moisture}, the \emph{skin temperature}, and the \emph{surface pressure}; and $273$ data corresponding to altitude profiles of \emph{temperature},
\emph{moisture}, and \emph{ozone} at $91$ model levels each.
An example of surface temperature is shown in Fig.~\ref{fig:map}. Data was provided by the official European Center for Medium-range Weather Forecasting (ECMWF) model, http://www.ecmwf.int/, on March 4th, 2008.

\begin{figure*}[t!]
\centering
\setlength{\tabcolsep}{1pt}
\begin{tabular}{cc}
\includegraphics[width = 8cm]{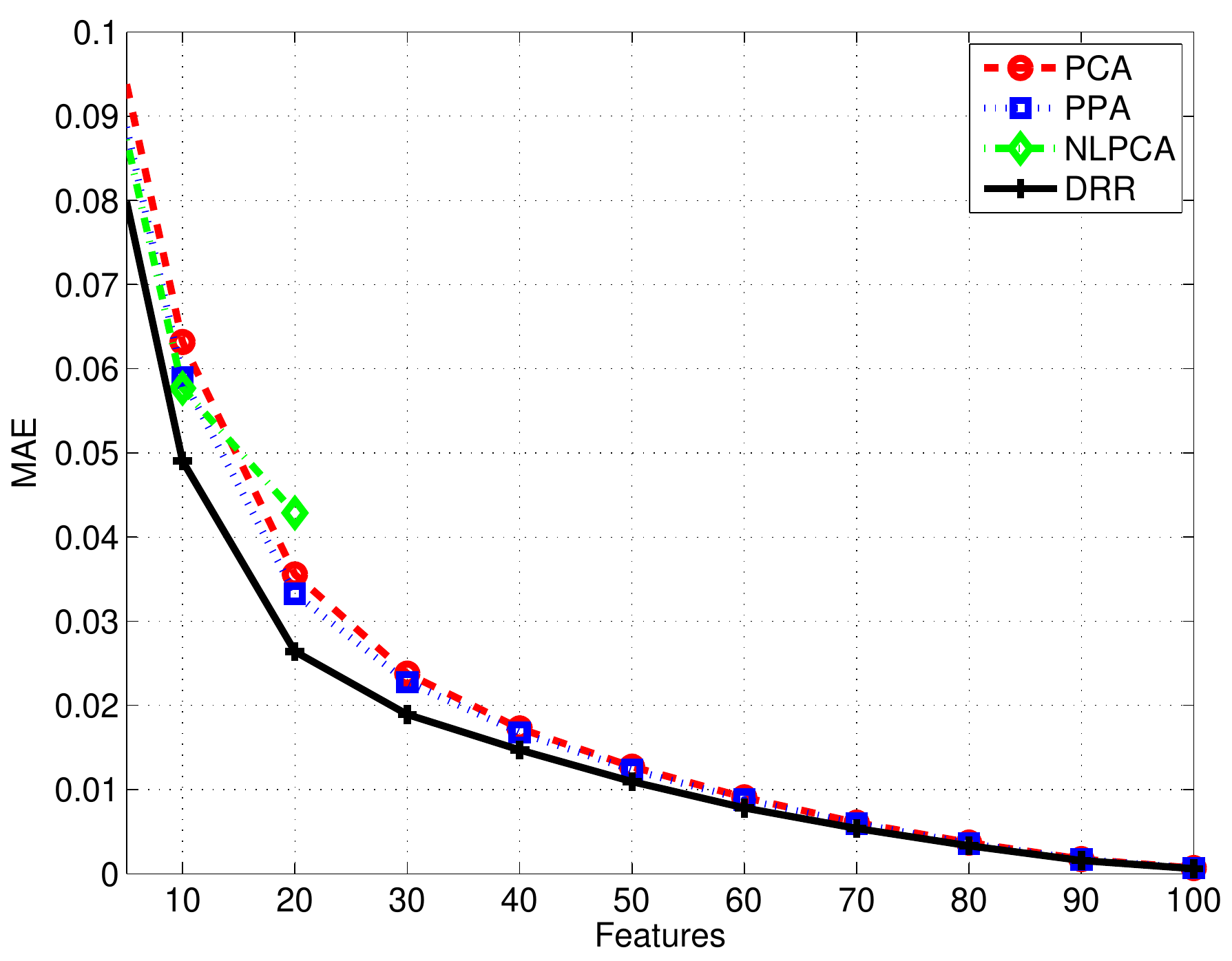} &
\includegraphics[width = 8cm]{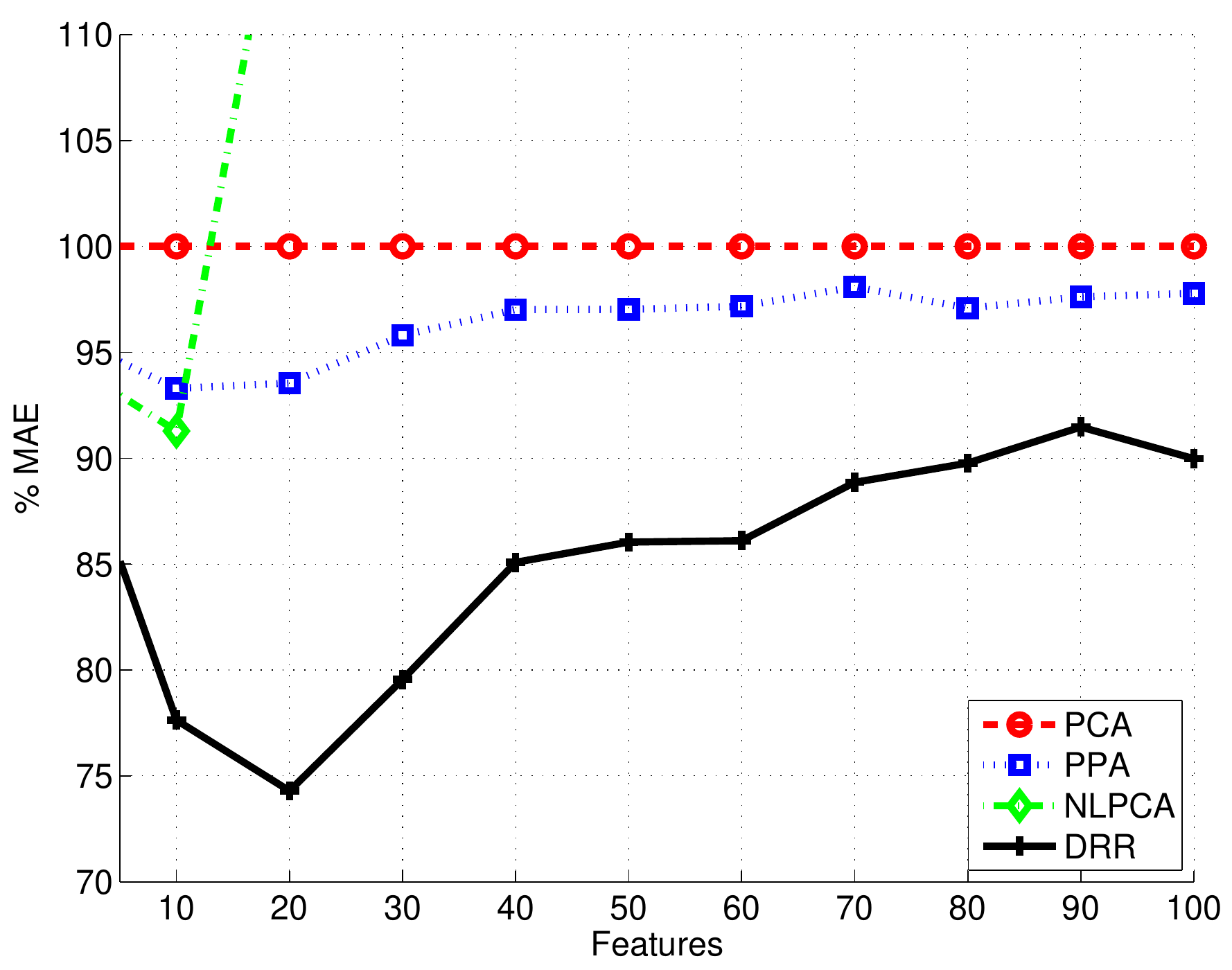}
\end{tabular}
\vspace{-0.25cm}
\caption{\label{fig:dr_error} Reconstruction error. Left: Absolute reconstruction error for different number of retained features obtained when using different DR methods on the first (just one) dataset. Right: Relative error (percentage) with regard to the error in PCA, mean and standard deviation have been obtained over the $23$ (all) datasets.
}
\end{figure*}

\subsubsection{Reconstruction accuracy}

In this experiment, we study the representation power of a small number of features extracted by DRR.
The 110 input features are processed with PCA~\cite{Jolliffe02}, PPA~\cite{Laparra12b,Laparra14a}, NLPCA~\cite{Kramer91,Scholz07} and the presented DRR method.
Here, the quality of the transformation is evaluated solely with the mean absolute error (MAE) in the input space between the original signal and the reconstructed with the most relevant coefficients retained. 
Figure~\ref{fig:dr_error} illustrates the effect of reconstructing the input data when using PCA, PPA, NLPCA and DRR for different numbers of components.
On the one hand, as reported in~\cite{Laparra14a}, the performance in PPA is similar or better than in NLPCA in reconstruction error. On the other hand, it is important to note that results in absolute and relative terms show that DRR clearly obtains less reconstruction error than PCA and PPA for an arbitrary number of features.

\subsubsection{Retrieval accuracy}

\begin{figure*}[t!]
  \centering
  \setlength{\tabcolsep}{1pt}
\begin{tabular}{cc}
\includegraphics[width = 8cm]{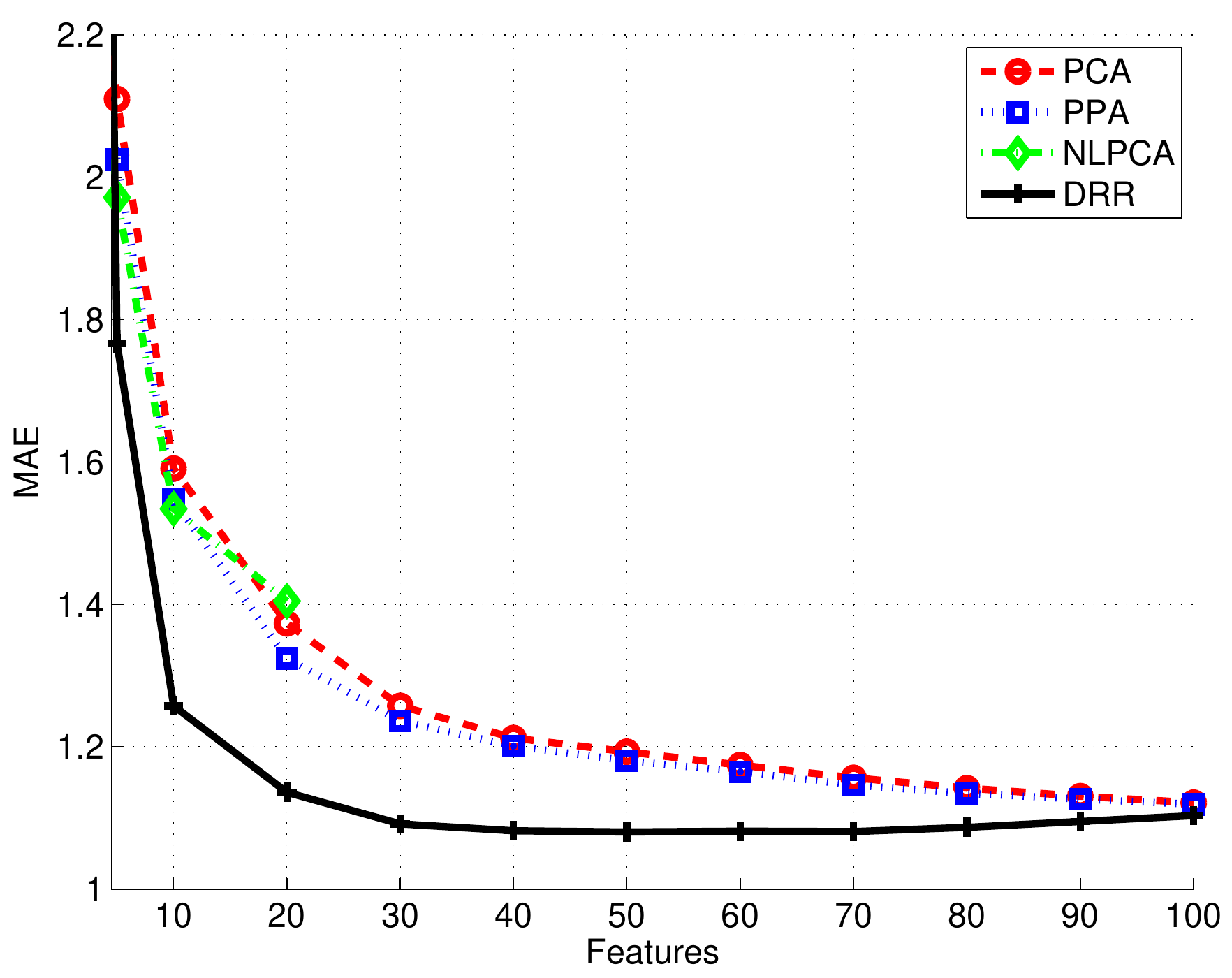} &
\includegraphics[width = 8cm]{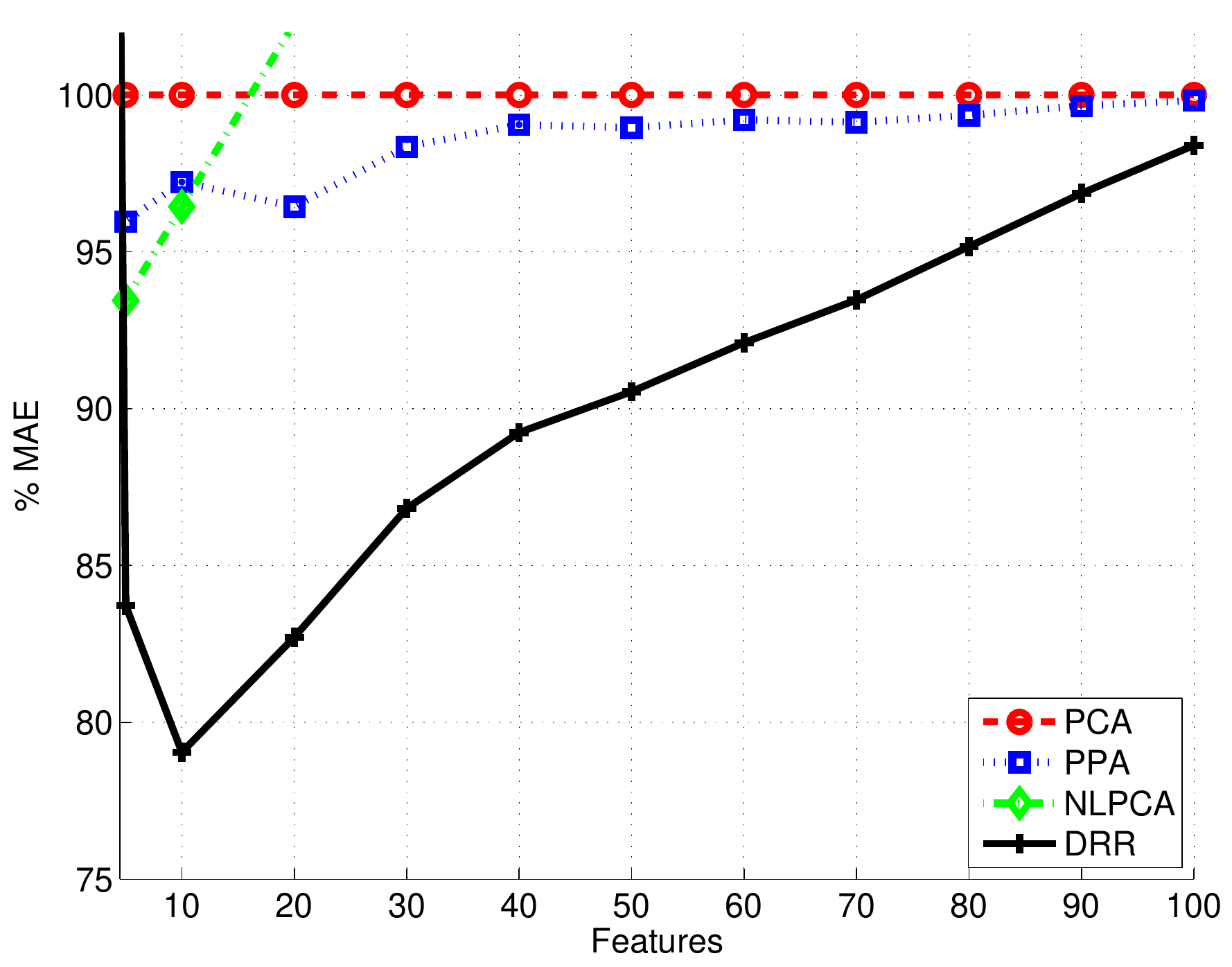}
\end{tabular}
  \caption{\label{fig:profiles} {Retrieval performance.} Accuracy of the parameter retrieval (MAE) with regard to the number of retained features.
  Results are given for different feature extraction (PCA, PPA, NLPCA, DRR) methods. Left: Absolute MAE for the first dataset. Right: Relative (to the PCA MAE in each dimension) results. Results for the remainder $23$ are similar.}
\end{figure*}

Figure~\ref{fig:profiles} illustrates the effect of using the features either from PCA, PPA or DRR for the retrieval of the physical parameters described before. We used linear regression in the features-to-parameters estimation. We plotted the mean absolute error (MAE) for different number of features. These plots show the effect of using different (linear and non-linear) dimensionality reduction methods for retrieval. Figure~\ref{fig:profiles} shows the results for the first dataset for illustration purposes (similar results were obtained for the remainder datasets). Note that using DRR features to estimate the features has clear benefits. For instance, using just the $20\%$ of the DRR features obtains the same accuracy as PCA when using all the components.

\subsubsection{Computational load}

Times for training and testing are shown in Table \ref{time_EUM} (same computer resources as before). In this experiment, we took $10000$ training and $10000$ test samples, and $d=110$. As in the previous experiment, NLPCA and DRR are the most expensive in training and test, respectively. 
In this experiment, however, times for DRR are notably higher due to the increase in dimensionality but mostly to the bigger training set. 

\begin{table}[h!]
\caption{Computational cost IASI Dataset}
\begin{center}
\begin{tabular}{|l|l|l|l|l|}
\hline
              & PCA & PPA & NLPCA & DRR \\ \hline \hline
Training time (sec) & 0.13 &  16 &   65389  &  14424  \\ \hline
Testing time (sec) & 0.01 &  0.3 &   1.3  &  1112  \\ \hline
\end{tabular}
\label{time_EUM}
\end{center}
\end{table}

\section{Conclusions}
\label{conclusions}

We introduced a novel unsupervised method for dimensionality reduction via the application of a multivariate nonlinear regression to approximate each projection from the higher variance scores. The method is shown to generalize PCA and to achieve more data compression (smaller MSE for a fixed number of retained components) and better features for prediction (less error in classification and regression problems) than competitive nonlinear methods like NLPCA and PPA. Besides, unlike other nonlinear dimensionality reduction methods, DRR is easy to apply, it has out-of-sample extension, it is invertible, and the learned transformation is volume-preserving. We focused on the challenging problems of spatial-spectral multispectral land cover classification, and atmospheric parameter retrieval from hyperspectral infrared sounding data. Extension of DRR to cope with multiset/output regression, as well as impact of the data dimensionality and noise sources, will be explored in the future. 

\section{Acknowledgments}

The authors wish to thank Tim Hultberg from the European Organisation for the Exploitation of Meteorological Satellites (EUMETSAT) in Darmstadt, Germany, for kindly providing the IASI datasets used in this paper.


\end{document}